\documentclass[preprint,12pt]{elsarticle}
\usepackage{booktabs}
\usepackage{amssymb}
\usepackage{amsmath}
\usepackage{hyperref} % Add this line
\usepackage{etex}
\usepackage[utf8]{inputenc}
\usepackage[T5]{fontenc}

\usepackage{xcolor}

\journal{arXiv}

\begin{document}

\begin{frontmatter}

%% Title, authors and addresses

%% use the tnoteref command within \title for footnotes;
%% use the tnotetext command for theassociated footnote;
%% use the fnref command within \author or \affiliation for footnotes;
%% use the fntext command for theassociated footnote;
%% use the corref command within \author for corresponding author footnotes;
%% use the cortext command for theassociated footnote;
%% use the ead command for the email address,
%% and the form \ead[url] for the home page:
%% \title{Title\tnoteref{label1}}
%% \tnotetext[label1]{}
%% \author{Name\corref{cor1}\fnref{label2}}
%% \ead{email address}
%% \ead[url]{home page}
%% \fntext[label2]{}
%% \cortext[cor1]{}
%% \affiliation{organization={},
%%             addressline={},
%%             city={},
%%             postcode={},
%%             state={},
%%             country={}}
%% \fntext[label3]{}

\title{FW-GAN: Frequency-Driven Handwriting Synthesis with Wave-Modulated MLP Generator} %% Article title

%% use optional labels to link authors explicitly to addresses:
% \author[label1,label2]{}
% \affiliation[label1]{organization={},
%             addressline={},
%             city={},
%             postcode={},
%             state={},
%             country={}}
%
% \affiliation[label2]{organization={},
%             addressline={},
%             city={},
%             postcode={},
%             state={},
%             country={}}

\author[UIT,VNU]{Huynh Tong Dang Khoa} %% Author name
\author[UIT,VNU]{Dang Hoai Nam}
\author[UIT,VNU]{Vo Nguyen Le Duy \corref{cor1}}

%% Author affiliation
\affiliation[UIT]{organization={University of Information Technology},
            city={Ho Chi Minh City},
            country={Vietnam}}
            
\affiliation[VNU]{organization={Vietnam National University},
            city={Ho Chi Minh City},
            country={Vietnam}}

%\affiliation[RIKEN]{organization={RIKEN Center for Advanced Intelligence Project},
%            city={Tokyo},
%            country={Japan}}
            
\cortext[cor1]{Corresponding author. E-mail: duyvnl@uit.edu.vn}      
%\affiliation[Corresponding]{country={Corresponding author. E-mail: duyvnl@uit.edu.vn}}

%% Abstract
\begin{abstract}
%% Text of abstract
Labeled handwriting data is often scarce, limiting the effectiveness of recognition systems that require diverse, style-consistent training samples. Handwriting synthesis offers a promising solution by generating artificial data to augment training. However, current methods face two major limitations. First, most are built on conventional convolutional architectures, which struggle to model long-range dependencies and complex stroke patterns. Second, they largely ignore the crucial role of frequency information, which is essential for capturing fine-grained stylistic and structural details in handwriting. To address these challenges, we propose FW-GAN, a one-shot handwriting synthesis framework that generates realistic, writer-consistent text from a single example. Our generator integrates a phase-aware Wave-MLP to better capture spatial relationships while preserving subtle stylistic cues. We further introduce a frequency-guided discriminator that leverages high-frequency components to enhance the authenticity detection of generated samples. Additionally, we introduce a novel Frequency Distribution Loss that aligns the frequency characteristics of synthetic and real handwriting, thereby enhancing visual fidelity. Experiments on Vietnamese and English handwriting datasets demonstrate that FW-GAN generates high-quality, style-consistent handwriting, making it a valuable tool for augmenting data in low-resource handwriting recognition (HTR) pipelines. Official implementation is available at \href{https://github.com/DAIR-Group/FW-GAN}{https://github.com/DAIR-Group/FW-GAN}

\end{abstract}

%%Graphical abstract
%\begin{graphicalabstract}
%\includegraphics[width=\textwidth]{architecture.png}
%\end{graphicalabstract}

%%Research highlights
%\begin{highlights}
%\item We propose WriteViT, a unified ViT-based architecture for one-shot handwriting synthesis, replacing CNN/CRNN modules with transformer-based components.
%\item Our model supports multi-scale generation and achieves state-of-the-art FID and KID scores on both English and Vietnamese datasets.
%\item We are among the first to apply handwriting synthesis to Vietnamese, a language with rich diacritics and structural complexity.
%\item The lightweight design enables deployment on low-resource platforms, while maintaining strong performance in both generation and recognition tasks.
%
%\end{highlights}

%% Keywords
\begin{keyword}
Handwritten text synthesis \sep Wavelet transform \sep One-shot learning \sep Vietnamese handwriting \sep Synthetic data
 
%% PACS codes here, in the form: \PACS code \sep code

%% MSC codes here, in the form: \MSC code \sep code
%% or \MSC[2008] code \sep code (2000 is the default)

\end{keyword}

\end{frontmatter}

%% Add \usepackage{lineno} before \begin{document} and uncomment 
%% following line to enable line numbers
%% \linenumbers

%% main text
%%

%% Use \section commands to start a section
\section{Introduction}\label{sec:introduction}
Despite rapid technological progress, handwritten documents continue to play an important role across a wide range of real-world applications because they often capture information in contexts where digital input is impractical or unavailable. In government institutions, handwritten forms are still widely used for signatures, annotations, and quick data collection, especially in low-digital or high-security environments. In cultural heritage and historical research, handwritten manuscripts, letters, and archival materials serve as primary records of linguistic, social, and political history, requiring precise preservation and digitization for scholarly study. Similarly, in education, handwritten exams and assignments remain the norm for evaluating students’ knowledge, creativity, and problem-solving skills in a natural and flexible medium. Beyond formal systems, handwriting is deeply tied to individual identity and expression, making it not only a functional tool but also a carrier of personality and nuance that typed text cannot fully convey. As a result, robust handwriting recognition (HTR) and synthesis technologies are essential to bridge analog and digital worlds in these domains. However, HTR systems face persistent challenges due to high variability in writing styles, differences in language structures, and varying image quality caused by scanning or photographic conditions \cite{HANDS-VNOnDB, kleber2013cvl, ICFHR2018, NabucoLatin}. These challenges are further amplified in low-resource languages, where acquiring large annotated datasets for training is costly, labor-intensive, or entirely impractical. 

Recent advances in deep learning, especially Transformer-based models, have yielded impressive performance gains in HTR. Yet, these models are heavily data-dependent and often fail to generalize well in data-scarce scenarios. Handwriting synthesis (HS) has emerged as a promising solution, offering automated generation of artificial samples to augment training datasets \cite{graves2013generating, bhunia2021handwriting}. Effective HS systems aim to generate lexically correct and stylistically consistent handwriting for arbitrary textual inputs. Achieving this, however, remains difficult due to the need to capture subtle visual cues such as stroke curvature, spacing irregularities, and writer-specific traits, especially when limited samples are available. These challenges are even more pronounced in complex scripts like Vietnamese, where tone marks and diacritic placement introduce additional orthographic variation.

While recent generative models, particularly those based on Generative Adversarial Networks (GANs), have improved synthesis realism, two core limitations still hinder performance. First, most existing approaches overlook the critical importance of frequency-domain information in handwriting synthesis. The F-Principle \cite{Zhi_2020FPrinciple} reveals that neural networks tend to learn low-frequency components first, leading to synthetic images that may preserve global structure but lack high-frequency details such as sharp stroke edges and pen pressure variations, which are essential for realistic handwriting. Second, many prior models adopt a CNN-based generator, which is inherently constrained by their limited receptive field. These convolutional structures emphasize local spatial features but fail to capture long-range dependencies and global frequency characteristics, both crucial for producing coherent and style-consistent handwriting across longer text sequences.

To address these challenges, we introduce FW-GAN, a frequency-aware, one-shot handwriting synthesis framework designed to produce realistic and writer-consistent handwritten text from a single reference. Our method enhances the generator with phase-aware Wave-MLP modules~\cite{Tang_2022WaveMlp}, which dynamically aggregate tokens, improving the synthesis of the model. We further introduce a novel Frequency Distribution Loss~\cite{Ni_2024FDL}, which explicitly supervises the generator to match the frequency characteristics of authentic handwriting. Additionally, we propose a high-frequency discriminator for detecting subtle artifacts and guiding the generator toward more faithful synthesis. Experiments on both English and Vietnamese handwriting datasets have verified the excellent performance of the proposed model. Its ability to augment low-resource training data positions it as a practical solution for enhancing HTR systems used in real-world expert applications. The main contributions to this paper are as follows:
\begin{itemize}
    \item We integrate Phase-Aware Wave-MLP modules that dynamically aggregate spatial tokens through trigonometric transformations, enabling better modeling of stylistic patterns while maintaining computational efficiency.
    \item We introduce a novel loss function that explicitly aligns the spectral characteristics between real and synthetic handwriting, ensuring generated samples preserve fine-grained frequency details essential for visual authenticity.
    \item We propose a frequency-guided discriminator alongside the standard spatial discriminator, where the high-frequency discriminator leverages wavelet decomposition to detect subtle artifacts that the spatial discriminator might miss, providing complementary adversarial supervision.
    \item Extensive experiments on English and Vietnamese handwritten datasets demonstrate that FW-GAN significantly outperforms the state-of-the-art.
\end{itemize}

\section{Related Work}\label{sec:related}

Handwriting text generation (HTG) encompasses a wide range of techniques aimed at producing synthetic handwritten content that resembles human writing in both content and style. These techniques can be broadly divided into two categories based on how handwriting is represented: online methods, which model the dynamic pen trajectory, and offline methods, which treat handwriting as a static visual artifact. Each paradigm offers unique strengths and faces specific limitations.

\subsection{Online Handwriting Generation}
Online HTG methods generate handwriting by simulating the pen’s motion over time. These techniques employ sequence-based models such as LSTMs~\cite{graves2013generating}, conditional variational recurrent networks~\cite{aksan2018deepwriting}, or temporal CNNs~\cite{aksan2018stcn} to predict a series of pen coordinates conditioned on the target text. The foundational work by Graves~\cite{graves2013generating} introduced this paradigm but did not incorporate style conditioning. Subsequent studies~\cite{aksan2018deepwriting, aksan2018stcn, kotani2020generating} addressed this gap by extracting style representations from exemplar images and integrating them into the sequence model.

Generative Adversarial Networks (GANs) have also been adapted for this modality - for instance, Ji et al.~\cite{ji2019generative} incorporated a discriminator to enhance pen trajectory realism. However, despite these advances, online methods face two fundamental obstacles. First, modeling long-range dependencies in pen motion remains difficult. Second, online data requires access to temporal stroke recordings, which are expensive to collect and unavailable for historical manuscripts. As a result, many recent works, including ours, have shifted focus toward the offline generation setting, where broader datasets and real-world applications are more accessible.

\subsection{Offline Handwriting Generation}

Offline HTG synthesizes handwriting directly in static images. Earlier efforts~\cite{wang2005combining, lin2007style, thomas2009synthetic, haines2016my} relied on manually segmented glyphs and heuristics for layout, ligature connections, and background blending. These approaches required substantial manual design and could not generalize to unseen characters or styles. Learning-based methods, particularly those using GANs, have largely replaced these hand-crafted pipelines~\cite{alonso2019adversarial, fogel2020scrabblegan, kang2020ganwriting, gan2021higan}. Alonso et al.\cite{alonso2019adversarial} introduced a GAN-based model that generated fixed-size word images from text embeddings. This was extended in ScrabbleGAN\cite{fogel2020scrabblegan} to variable-length synthesis using a patch-based character assembly approach. Subsequent works~\cite{kang2020ganwriting, gan2021higan, mattick2021smartpatch} have adopted similar strategies, often conditioning generation on text encoded as character-level one-hot vectors. Other approaches~\cite{luo2022slogan, krishnan2021textstylebrush} perform direct image-to-image style transfer using typeface-rendered input images. Furthermore, style conditioning in modern offline HTG models varies in granularity, ranging from paragraph-level references~\cite{davis2020text} to word-level or even single-word exemplars~\cite{gan2021higan, luo2022slogan}. In general, richer style inputs lead to better synthesis fidelity~\cite{krishnan2021textstylebrush}. Despite their strong visual performance, most existing methods rely on convolutional backbones derived from BigGAN~\cite{brock2019biggan}, which are limited in capturing global structural patterns due to their restricted receptive fields. As a result, these models tend to focus heavily on local features, which hinders their ability to learn the overall handwriting structure and may lead to suboptimal style representation.

However, combining CNNs and Transformers to jointly model global and local dependencies introduces significant computational overhead~\cite{bhunia2021handwriting, pippi2023handwritten}, which is particularly problematic due to the extensive training requirements of GANs. In the computer vision field, while several works have explored hybrid architectures that integrate Transformers with CNNs or MLPs~\cite{tolstikhin2021mlpmixer, ding2021repmlp, Guo_2022BeyondSelfAttention, bhunia2021handwriting, pippi2023handwritten}, these models often suffer from high complexity due to their large number of parameters. On the other hand, recent progress in MLP-based networks~\cite{melaskyriazi2021need, Tang_2022WaveMlp, liu2021swin, ho2019axial} shows that they can achieve competitive performance compared to CNNs while being more lightweight, making them an attractive option for efficient generation. Motivated by this, we explore the combination of CNN and MLP modules in the generation process, aiming to develop a handwriting generator that balances high-quality output with computational efficiency.

In addition to architectural limitations, many existing approaches overlook frequency-domain characteristics. The Frequency Principle (F-Principle)~\cite{Zhi_2020FPrinciple} suggests that during training, neural networks tend to learn low-frequency (smooth) components of the target function before fitting high-frequency (detailed) components. This principle has been theoretically supported in various contexts, including infinite data regimes~\cite{luo2019theory}, wide neural networks in the Neural Tangent Kernel (NTK) regime~\cite{jacot2018ntk}, and finite sample settings~\cite{luo2020theory, zhang2019explicitizing, bordelon2020spectrum, cao2019towards, basri2019convergence, yang2019fine}. Additionally, E et al.~\cite{e2019frequency} demonstrate that this behavior can naturally arise from the integral formulation of network training. Based on this insight, analyzing the frequency components of real and generated handwriting images may help uncover key differences and guide further improvements in model design.

Building upon these insights, we enhance the generator by integrating Wave-MLP modules into the convolutional blocks of the BigGAN backbone, commonly adopted in several state-of-the-art methods~\cite{gan2021higan, gan2022higan+, bhunia2021handwriting, pippi2023handwritten}. These modules dynamically aggregate spatial information through trigonometric transformations, improving synthesis quality while maintaining a compact model size. To further enhance realism, we introduce a Frequency Distribution Loss (FDL) that aligns the spectral statistics of generated handwriting with those of real samples. Additionally, we deploy a frequency-guided discriminator alongside the standard spatial discriminator, where the high-frequency discriminator leverages wavelet decomposition to detect subtle artifacts that the spatial discriminator might miss, providing complementary adversarial supervision across both spatial and frequency domains. We adopt the framework similar to HiGAN ~\cite{gan2021higan} as a practical baseline due to its stable training behavior and compatibility with a BigGAN-style generator to develop our proposed methods. However, our proposed architectural modifications and training objectives are independent of this framework and can be readily incorporated into alternative generative models. An overview of the proposed framework is illustrated in Figure~\ref{fig:arch}.

\section{Proposed Approach}\label{sec:proposal}
\subsection{Problem Formulation}\label{subsec:formulation}
Given a single handwritten word image $\mathbf{x}$ authored by a target writer $\mathbf{w}$, which has a calligraphic style $\mathbf{z}$, our goal is to synthesize new handwritten images that preserve the unique calligraphic style of writer $\mathbf{w}$. We define a query set $A = \{a_k\}_{k=1}^Q$, where each $a_k$ is a word of arbitrary length composed from a general character set. Each query word $a_k$ is represented as a character sequence $a_k = [a_{k,1}, \ldots, a_{k,L_k}]$, where $L_k$ is the length of the $k$-th word. To generate a corresponding handwritten image $\hat{\mathbf{x}}_k$ in the style of $\mathbf{x}$, we train a generator $G$ that maps a character sequence and a style vector to an image:
\begin{equation}
\hat{\mathbf{x}}_k = G(a_k, \mathbf{z}).
\end{equation}

There are two ways of obtaining the style vector $\mathbf{z}$: it can either be sampled from a standard normal distribution $\mathcal{N}(0, \mathbf{I})$ or extracted directly from the reference image $\mathbf{x}$ using a style encoder $E$ by $\mathbf{z} = E(\mathbf{x})$. This flexibility enables the generator to reproduce the specific handwriting style of the target writer.

\begin{figure*}[!t]
	\footnotesize
	\centering
	\includegraphics[width=1\linewidth]{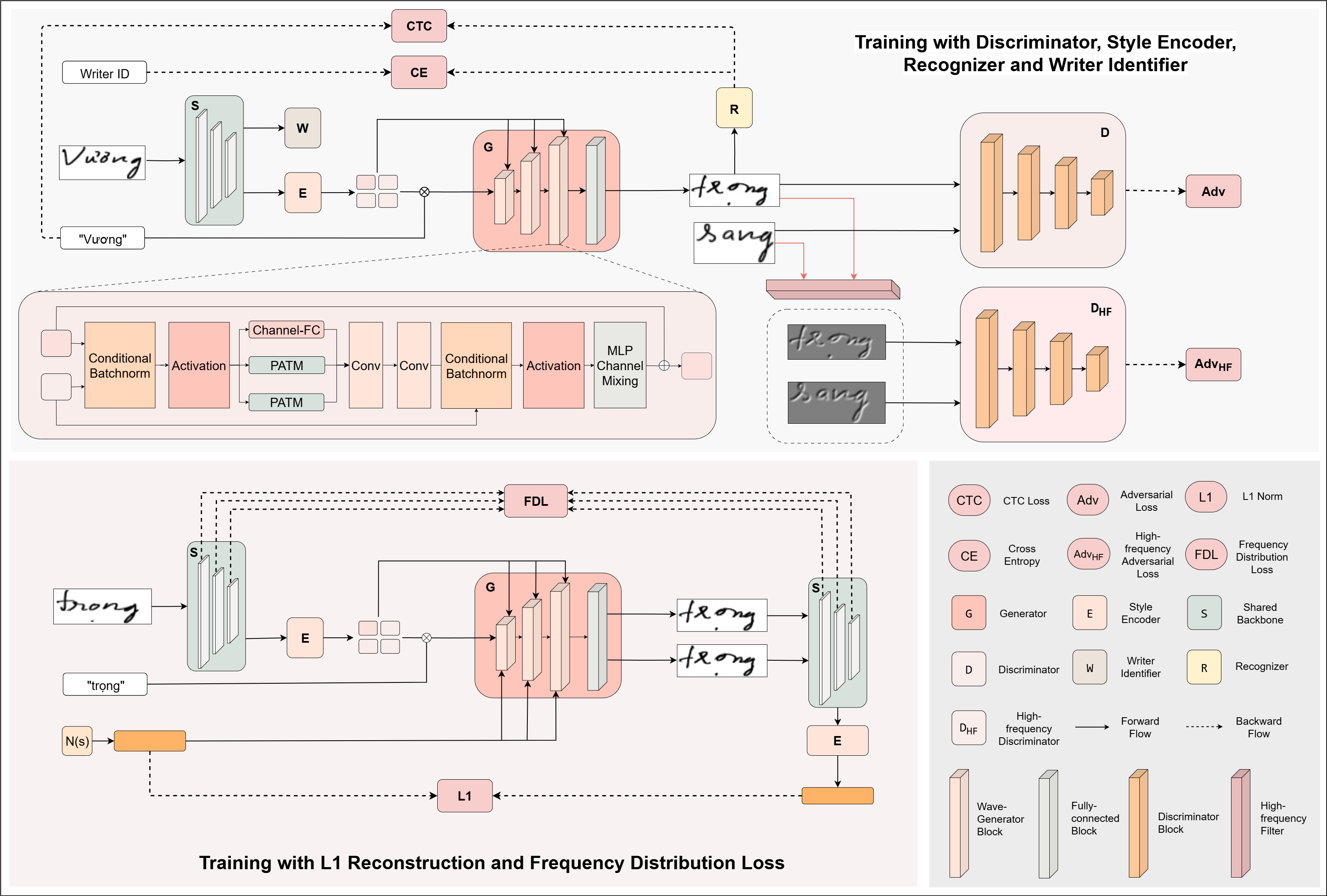}
	\caption{Overview of the proposed FW-GAN}
	\label{fig:arch}
	\footnotesize
\end{figure*}

\subsection{Overall Architecture}
Our model features a hierarchical generator composed of WaveGBlocks, which utilize phase-aware Wave-MLP layers to effectively capture spatial patterns and preserve the sequential structure of character layouts, supporting the generation of variable-length handwritten text. The generator is conditioned on both character-level content embeddings and writer-specific style features, enabling personalized and flexible synthesis. To enhance realism and preserve structural coherence, we deploy a dual discriminator architecture where a standard spatial discriminator and a frequency-guided discriminator work in parallel, with the latter leveraging high-frequency wavelet components to detect subtle artifacts that spatial analysis might miss. The system also includes a recognizer to enforce content accuracy, a style encoder to extract personalized writing styles, and a writer identifier to maintain writer-specific characteristics during training. All components are trained together in a unified framework, where the generator learns to produce handwriting that is not only legible and accurate but also visually consistent with the target style while maintaining authentic frequency characteristics.

\begin{figure*}[!t]
	\footnotesize
	\centering
	\includegraphics[width=1\linewidth]{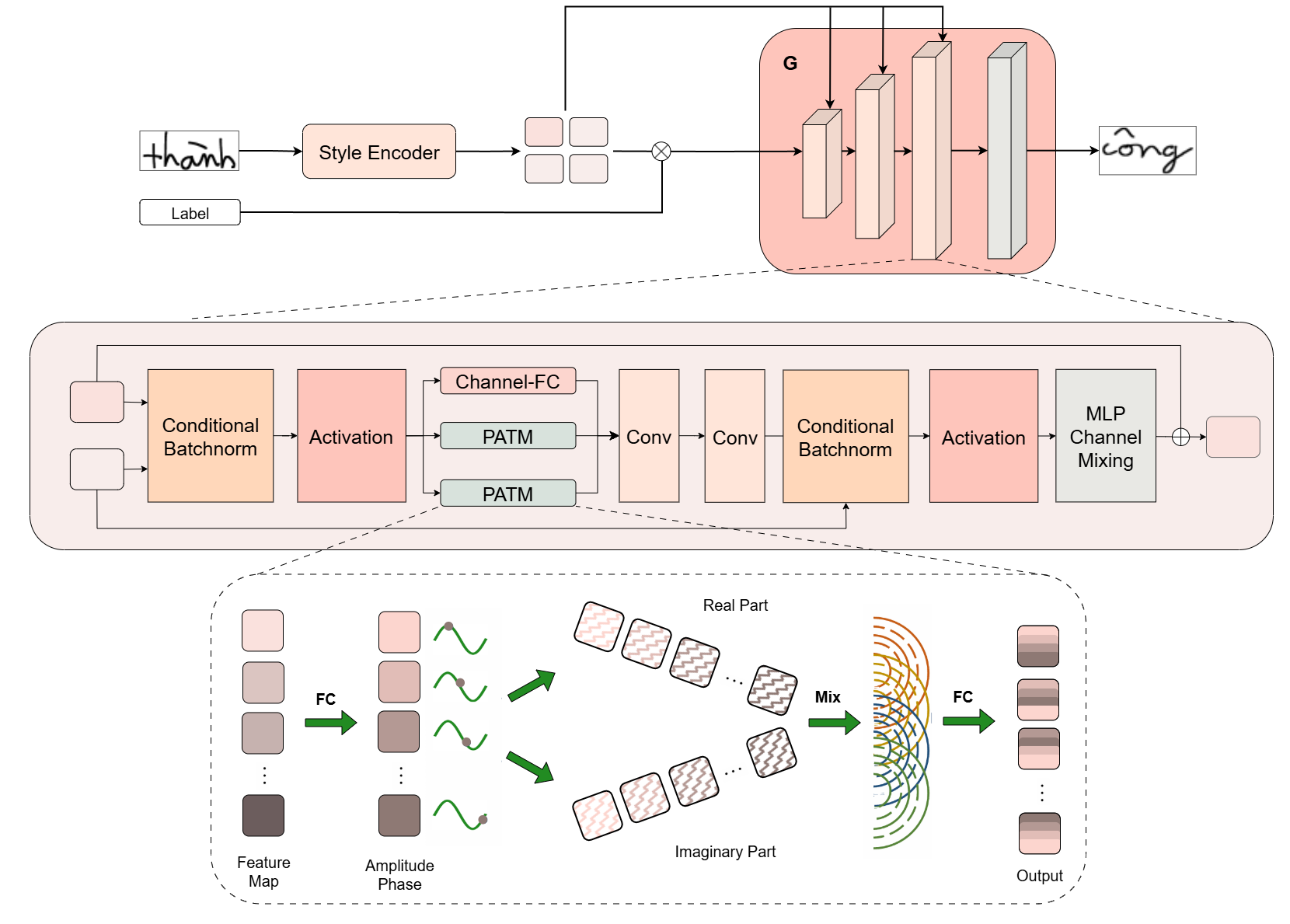}
	\caption{Overview of the proposed WaveGBlock}
	\label{fig:wavegblock}
	\footnotesize
\end{figure*}

\subsubsection{Hierarchical Generator with Wave-MLP}
\label{subsubsec:generator}

Our handwriting synthesis framework employs a hierarchical generator, jointly conditioned on a writer-specific latent style vector $\mathbf{z} \in \mathbb{R}^d$ and a symbolic character sequence $\mathbf{y} \in \{0,1\}^{n \times L}$, where $n$ is the vocabulary size and $L$ is the sequence length. The style vector $\mathbf{z}$ is divided into multiple segments: the first segment is fused with the character input $\mathbf{y}$ (via elementwise multiplication) to form a unified latent representation. This representation is then projected and reshaped to initialize the feature map $\mathbf{F}_0 \in \mathbb{R}^{C_0 \times H_0 \times W_0}$. The remaining segments of $\mathbf{z}$ are gradually injected into subsequent Wave-Modulated Generator's Blocks (WaveGBlocks) using conditional batch normalization, allowing the style information to guide feature transformations at each stage. The feature map is progressively upsampled and refined through the hierarchy of WaveGBlocks, ultimately producing a variable-width handwriting image that accurately reflects both the content and the writer’s style.

Each WaveGBlock introduces a new mechanism, Wave-MLP for feature interaction, inspired by wave-based representations in quantum mechanics. In this framework, entities such as electrons and photons are described by wave functions containing both amplitude and phase components (e.g., de Broglie waves). Drawing from this idea, we define tokens as spatial units in the intermediate feature maps, analogous to tokens in sequence models but representing local image regions. These tokens are modeled as waves with amplitude and phase that can dynamically interact. The amplitude represents the token’s real-valued content, while the phase encodes its relational role and direction, forming a complex-valued signal through interaction with fixed MLP weights. Wave-MLP incorporates a Phase-Aware Token Mixing (PATM) module to dynamically estimate the phase for each token, as semantic meaning can vary with input. In Wave-MLP, token is represented as a Wave\textasciitilde$\mathbf{f}_j$ with amplitude and phase information:
\begin{equation}
\label{eq:wave_signal}
\tilde{\mathbf{f}}_j = |\mathbf{f}_j| \odot e^{i\theta_j}, ~j = 1, 2, 3,\dots,n,
\end{equation}

where $|\mathbf{f}_j|$ is the real-valued amplitude, $\theta_j$ is the learned phase parameter, and $\odot$ denotes elementwise multiplication. 

The Wave-MLP framework, illustrated in Figure~\ref{fig:wavegblock}, comprises two main modules: (1) Token Mixing, which aggregates tokens using amplitude and phase information; and (2) Channel MLP Mixing, which extracts intra-token features through two channel-wise fully connected layers with nonlinear activations. Token Mixing integrates a channel-wise fully connected layer (Channel-FC) with two Phase-Aware Token Mixing (PATM) modules operating along spatial dimensions. PATM plays the central role by modeling each token as a waveform and enabling interactions via both amplitude and phase. As shown in Eq.~\ref{eq:wave_signal}, these wave-like tokens are initially represented in the complex domain. To integrate them into an MLP-style architecture, we can apply Euler’s formula to unfold the complex expression into real and imaginary parts:
\begin{align}
\tilde{z}_j = |z_j| \odot \cos \theta_j + i |z_j| \odot \sin \theta_j, \quad j = 1, 2, \cdots, n.
\end{align}

This formulation expresses each complex token as a pair of real-valued vectors, representing its real and imaginary components. These tokens $\tilde{z}_j$ are then aggregated using the Token-FC operation:
\begin{equation}
\label{eq:Token-FC}
\tilde{o}_j = \text{Token-FC}(\tilde{F}, W^t)_j, \quad j = 1, 2, \cdots, n,
\end{equation}
where $\tilde{F} = [\tilde{f}_1, \tilde{f}_2, \cdots, \tilde{f}n]$ represents the set of all wave-like tokens. In this step, tokens interact with one another by taking both amplitude and phase into account. The result $\tilde{o_j}$ is the complex-valued output of the aggregated token. Following common quantum measurement approaches~\cite{Braginsky1992_quantummeasurement, Jacobs2006_quantummeasurement}, we can project this complex output into the real domain by summing its real and imaginary parts with learnable weights:
\begin{equation}
o_j = \sum_k W^t_{jk} f_k \odot \cos \theta_k + W^i_{jk} f_k \odot \sin \theta_k, \quad
j = 1, 2, \cdots, n,
\label{eq:wave_output}
\end{equation}
where $W^t$ and $W^i$ are trainable weights. Here, the phase $\theta_k$ dynamically modulates the aggregation process based on the semantic properties of the input.

Two Phase-Aware Token Mixing (PATM) modules operate along the height and width dimensions to capture spatial dependencies in both directions. These branches are combined through a reweighting mechanism inspired by CycleMLP, which adaptively fuses directional information. By representing each token as a waveform and modulating it with learned phase information, our generator achieves both long-range spatial coherence and fine-grained stylistic control, enabling high-quality handwriting synthesis that remains faithful to the input text and the writer’s unique style.

\begin{figure*}[!t]
	\footnotesize
	\centering
	\includegraphics[width=1\linewidth]{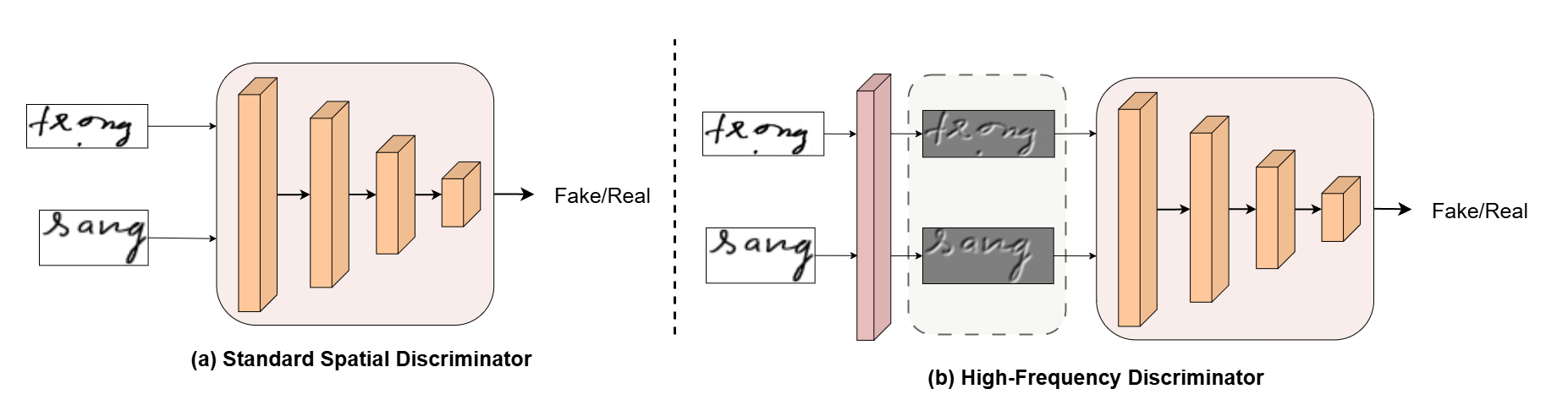}
	\caption{Illustration of the Dual Discriminator Architecture.}
	\label{fig:disc-arch}
	\footnotesize
\end{figure*}

\subsubsection{Dual Discriminator Architecture} \label{subsubsec:discriminator}

To enhance the discriminative capability and address artifacts that may not be readily apparent in the spatial domain alone, we employ a dual discriminator architecture consisting of two complementary components: a standard spatial discriminator $D$ and a high-frequency discriminator $D_{HF}$, as illustrated in Figure~\ref{fig:disc-arch}.

\paragraph{Standard Spatial Discriminator}
The standard discriminator $D$ follows the conventional adversarial framework, processing input handwriting images through a sequence of convolutional blocks. Each block consists of convolutional layers with spectral normalization, batch normalization, and ReLU activations, progressively reducing spatial resolution while increasing feature depth. This hierarchical feature extraction captures local textures, character shapes, and structural patterns at multiple scales. The final feature maps are globally averaged and passed through a linear layer to produce a spatial realism score $D(x)$, which reflects the plausibility of the image in the spatial domain. 

\begin{figure*}[!t]
	\footnotesize
	\centering
	\includegraphics[width=1\linewidth]{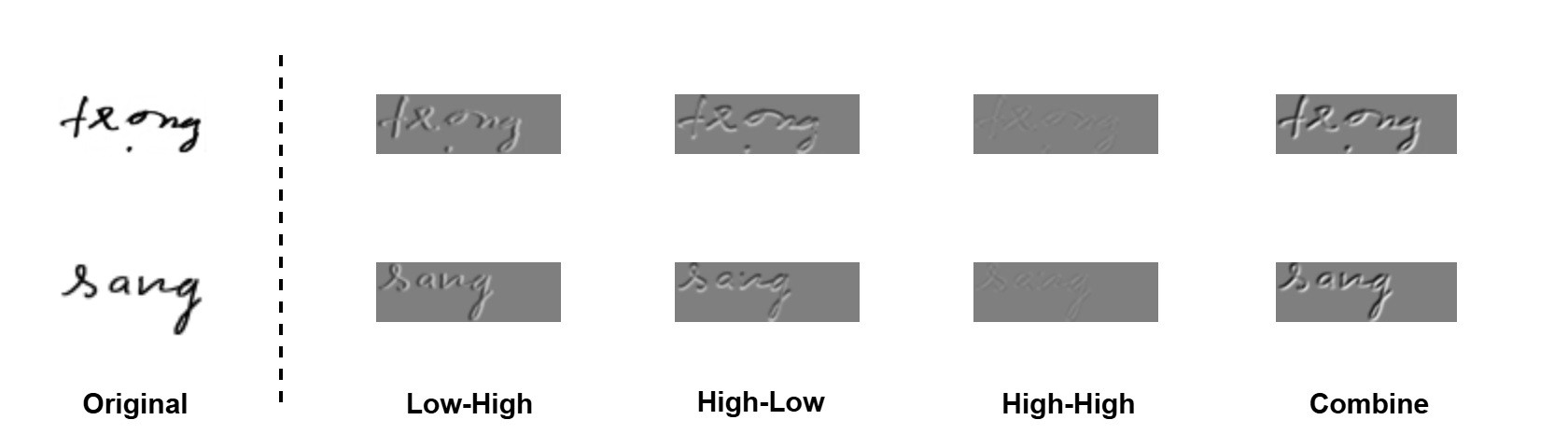}
	\caption{Visualization of the transformed frequency components of handwriting images}
	\label{fig:high-freq-img}
	\footnotesize
\end{figure*}

\paragraph{High-Frequency Discriminator} 
To complement spatial analysis and target high-frequency artifacts common in generated handwriting, we introduce a dedicated high-frequency discriminator $D_{HF}$. This discriminator operates on frequency-decomposed representations of the input images rather than on raw pixel intensities. It employs a two-dimensional Haar wavelet transform implemented via grouped convolutions with fixed Haar wavelet kernels. The wavelet decomposition for high-frequency components is formally expressed as:
\begin{equation}
\mathcal{W}(x) = \{W_{LH}, W_{HL}, W_{HH}\},
\end{equation}
where the decomposition yields three sub-bands: $W_{LH}$ (low--high), $W_{HL}$ (high--low), and $W_{HH}$ (high--high). Figure~\ref{fig:high-freq-img} shows that the high-frequency sub-bands ($W_{LH}, W_{HL}, W_{HH}$) capture detail coefficients corresponding to sharp edges, stroke boundaries, and fine contour variations that are crucial for authentic handwriting synthesis. The Haar wavelet basis functions are defined as:
\begin{equation}
h_L = \frac{1}{\sqrt{2}}[1, 1], \quad h_H = \frac{1}{\sqrt{2}}[1, -1].
\end{equation}

The 2D wavelet filters are constructed as tensor products of the 1D basis functions: $W_{LH} = h_L^T \otimes h_H$, $W_{HL} = h_H^T \otimes h_L$, and $W_{HH} = h_H^T \otimes h_H$, where $\otimes$ denotes the tensor product operation. This discriminator focuses on the detail coefficients that capture fine-grained textures and structural patterns characteristic of authentic handwriting. The high-frequency representation is obtained by aggregating the detail sub-bands:
\begin{equation}
\mathcal{H}(x) = W_{LH} + W_{HL} + W_{HH}.
\end{equation}

This aggregated high-frequency representation $\mathcal{H}(x)$ is then processed through the same convolutional architecture as the standard discriminator, yielding a frequency-domain realism score $D_{HF}(\mathcal{H}(x))$ that evaluates the authenticity of fine-scale features and textural details in the generated handwriting samples.

\subsubsection{Recognizer}
\label{subsec:text_recognizer}

The recognizer module $R$ is responsible for inferring the textual content $y$ from handwriting images. It is trained solely on real, annotated handwriting data and does not receive any supervision from synthetic examples. During generation, $R$ acts as a semantic regulator by enforcing textual accuracy in the output of the generator $G$, ensuring that the produced handwriting is not only visually coherent but also linguistically faithful to the intended input text.

\subsubsection{Style Encoder and Writer Identifier}
\label{subsec:style_writer_modules}

The style encoder $E$ extracts stylistic attributes from handwritten word images into a fixed-size latent vector $\mathbf{s}$ that conditions the generator $G$. The writer identifier $W$ predicts author identity from handwriting images, providing explicit stylistic supervision during training. Both modules share a fundamental objective: extracting style-relevant features while ignoring textual content. This similarity motivates a shared backbone architecture that processes input images through hierarchical convolutional layers with residual connections. The shared feature extraction reduces computational overhead while enabling both tasks to benefit from common visual representations.

Each module employs a task-specific head after the shared backbone. The style encoder uses a variational approach to model the style distribution, while the writer identifier performs classification over writer identities. Both incorporate sequence-aware processing to handle variable-length inputs. This design leverages multi-task learning benefits, where the complementary supervision signals from style extraction and writer classification mutually improve the quality of learned style representations.

\subsection{Objective Functions}

Before training, our framework requires a dataset that provides multiple forms of supervision: visual samples of handwriting ($\mathbf{X}$), their associated labels ($\mathbf{Y}$), and their corresponding author identifiers ($\mathbf{W}$). While these annotated samples serve as foundational data, our generative model must generalize beyond them, especially for synthesizing unseen or out-of-vocabulary (OOV) text. To address this, we draw textual inputs from a broader corpus, enabling the system to handle arbitrary word sequences during training by sampling a text sequence $\tilde{y}$ from the open corpus $\mathcal{C}$. The complete training strategy, including architecture and loss design, is outlined in Figure~\ref{fig:arch} and detailed below.

\subsubsection{Adversarial Loss}

We adopt the standard Generative Adversarial Network (GAN) framework, where the discriminator \( D \) learns to differentiate between real handwriting samples and those synthesized by the generator $G$. This adversarial setup encourages $G$ to produce more visually realistic handwriting. For stability and effectiveness, we employ the widely used hinge loss formulation~\cite{lim2017geometric}, defined as:
\begin{equation}
\mathcal{L}_{adv} = \mathbb{E}_{x \sim \mathbf{X}} \left[ \max(0, 1 - D(x)) \right] + \mathbb{E}_{\tilde{y} \sim \mathcal{C},\ z} \left[ \max(0, 1 + D(G(\tilde{y}, z))) \right],
\end{equation}
where the style feature $\mathbf{z}$ is obtained either by (1) sampling from a prior normal distribution \( \mathcal{N}(0,1) \), or (2) extracting it from a reference image \( x \) via the encoder, i.e., $\mathbf{z} = E(\mathbf{x})$. It is important to note that the adversarial loss focuses solely on improving the visual realism of the generated images. It does not enforce the preservation of textual content or the fidelity of calligraphic style, which are addressed by additional objectives in our framework.

\subsubsection{Text Recognition Loss}

To ensure the synthesized handwriting conveys the intended textual content, we introduce a handwriting recognizer $R$ to guide the generator $G$. While the adversarial loss promotes realism, it does not guarantee that the generated images preserve textual accuracy. The recognizer $R$ addresses this gap by explicitly aligning image content with textual targets. It is first trained in a supervised manner using real handwriting samples $x \in \mathbf{X}$ and their corresponding transcriptions $y \in \mathbf{Y}$. We employ the Connectionist Temporal Classification (CTC) loss~\cite{Graves2006ctc} to accommodate unsegmented sequence learning, defined as:
\begin{equation}
\mathcal{L}_{R}^{D} = \mathbb{E}_{x, y} \left[ \mathcal{L}_{\text{CTC}}(R(x), y) \right],
\end{equation}
where $R(x)$ is the sequence of predicted character probabilities for image $x$, and $y$ is the ground-truth label.

Once $R$ is trained, its parameters are frozen, and it serves as a perceptual guide to enforce textual fidelity in generated images. Specifically, for any sampled text $\tilde{y} \in \mathcal{C}$ and style feature $z$, the generator $G$ is encouraged to produce an image $G(\tilde{y}, z)$ such that $R(G(\tilde{y}, z))$ accurately predicts $\tilde{y}$. This leads to the generator-side recognition loss:
\begin{equation}
\mathcal{L}^{G}_{R} = \mathbb{E}_{\tilde{y}, z} \left[ \mathcal{L}_{\text{CTC}}(R(G(\tilde{y}, z)), \tilde{y}) \right],
\end{equation}
which is backpropagated through $G$ to improve its capacity to render syntactically accurate handwriting. This design ensures that generated samples not only appear realistic but also encode the correct textual information.

\subsubsection{Writer Identification Loss}

While handwriting appearance is often visually diverse, it remains stylistically consistent for a given individual. This property enables us to leverage writer identity as a proxy for style supervision. Since we lack explicit labels for fine-grained style attributes (e.g., stroke thickness, slant, character curvature), we introduce a writer classifier $W$ to capture these underlying style representations. The identifier $W$ is trained to predict the writer identity from real handwriting samples $x \in \mathbf{X}$ with associated writer labels $w \in \mathbf{W}$. This is achieved using a standard cross-entropy classification objective:
\begin{equation}
\mathcal{L}_{W}^{D} = \mathbb{E}_{x, w} \left[ - \log p(w \mid W(x)) \right],
\end{equation}
which encourages $W$ to learn writer-discriminative features that implicitly reflect style.

Once trained, $W$ is held fixed and used to supervise the generator $G$ via a feedback signal that enforces stylistic fidelity. Given a reference image $x$, we extract the style code $z = E(x)$ and synthesize a new image $G(\tilde{y}, z)$ with arbitrary content $\tilde{y} \in \mathcal{C}$. We then require the generated sample to be classified by $W$ as having originated from the same writer as the reference:
\begin{equation}
\mathcal{L}^{G}_{W} = \mathbb{E}_{x, w, \tilde{y}} \left[ - \log p(w \mid W(G(\tilde{y}, E(x)))) \right].
\end{equation}

This feedback loop ensures that the generator adheres to the stylistic identity encoded in the reference, even when synthesizing unseen textual inputs. It is important to note that the writer classifier $W$ is only trained on the training set and may not generalize to authors outside this domain (e.g., unseen writers in the test set).

\subsubsection{Style Reconstruction Loss}

To ensure that the generator genuinely incorporates the intended style feature during synthesis, we introduce a reconstruction constraint that enforces consistency between the input style and the one inferred back from the generated image, similar to the method proposed in \cite{chen2016infogan}. This is essential for learning an invertible mapping between the latent style space and visual handwriting appearance. Given a style vector $z \sim \mathcal{N}(0,1)$ and an arbitrary text $\tilde{y} \in \mathcal{C}$, the generator $G$ synthesizes an image $G(\tilde{y}, z)$. We then re-encode this image using the style encoder $E$ and require that the recovered style closely match the original:
\begin{equation}
\mathcal{L}_{style} = \mathbb{E}_{\tilde{y}, z} \left[ \left\| z - E(G(\tilde{y}, z)) \right\|_1 \right].
\end{equation}

This self-consistency loss encourages the model to meaningfully utilize the style vector $z$ during generation. It also promotes diversity in the output space and helps prevent the generator from collapsing into a limited set of styles.

\subsubsection{KL-Divergence Loss}

To enable stochastic style sampling during inference, we enforce the learned style space to follow a predefined prior distribution. This is achieved by applying a Kullback-Leibler (KL) divergence penalty between the distribution of encoded style vectors and a standard normal prior. Given a real handwriting image $x \in \mathbf{X}$, we compute the KL divergence between the posterior distribution $E(x)$ and the prior $\mathcal{N}(0,1)$:
\begin{equation}
\mathcal{L}_{kl} = \mathbb{E}_{x} \left[ D_{KL}(E(x) \, \| \, \mathcal{N}(0, 1)) \right].
\end{equation}

This loss acts as a regularizer that encourages the latent style space to remain continuous, smooth, and sampleable at test time. It is commonly used in variational frameworks and has proven effective in style transfer tasks~\cite{zhu2017multimodal, lee2018diverse}.

\subsubsection{Frequency Distribution Loss}

Standard image synthesis losses, such as pixel-wise $\ell_1$ or perceptual losses based on spatial features (e.g., Gram or contextual loss), typically assume spatial alignment between predicted and reference images. However, this assumption often breaks down in handwriting synthesis, where stylistic attributes, such as character shape, stroke curvature, and slant, can vary independently of exact spatial positioning. Additionally, content mismatches and varying text lengths further contribute to spatial misalignments between generated and real samples.

To address these issues, we adopt the Frequency Distribution Loss (FDL)~\cite{Ni_2024FDL}, applied on \textit{high-level feature representations}. The key idea is to treat an image as a set of semantic features and to measure similarity between distributions of these features while ignoring their spatial locations. This encourages stylistic consistency while allowing minor spatial variations. Specifically, we extract deep features from a fixed network $\Phi$ (e.g., the $l$-th layer of the shared backbone of writer identifier and style encoder), treating the outputs as unordered sets of feature vectors:
\begin{equation}
A = \{ a_1, a_2, \dots, a_N \} = \Phi^l(x), \quad
B = \{ b_1, b_2, \dots, b_N \} = \Phi^l(G(y, E(x))),
\end{equation}
where $x$ is a real handwriting image, and $G(y, E(x))$ is the reconstruction generated from the ground-truth content $y$ and the encoded style $E(x)$.

We apply the Discrete Fourier Transform (DFT) to the spatial axes of each feature map in $A$ and $B$, decomposing them into frequency domain representations. Specifically, each feature vector is transformed into a complex-valued representation from which we derive amplitude and phase components. Let $\mathcal{A}(A)$ and $\mathcal{P}(A)$ denote the sets of amplitudes and phases computed from $A$, and similarly $\mathcal{A}(B)$ and $\mathcal{P}(B)$ from $B$. To compare these distributions while accounting for local style variations and global structure, we apply the Sliced Wasserstein Distance (SWD) separately on amplitudes and phases:
\begin{equation}
\mathcal{L}_{\text{FDL}} = \mathrm{SW}(\mathcal{A}(A), \mathcal{A}(B)) + \lambda \cdot \mathrm{SW}(\mathcal{P}(A), \mathcal{P}(B)),
\end{equation}
where $\mathrm{SW}(\cdot, \cdot)$ denotes the SWD between two empirical distributions. The hyperparameter $\lambda$ balances the contribution of amplitude and phase statistics.

\subsubsection{High-Frequency Adversarial Loss}
While the standard adversarial loss encourages global visual realism, handwriting synthesis also depends on fine-grained attributes such as stroke texture, pen pressure variations, and sharp character edges. To explicitly enforce realism in these high-frequency details, we leverage the high-frequency discriminator $D_{HF}$ described in~\ref{subsubsec:discriminator}, which operates on the three high-frequency sub-bands (LH, HL, HH) obtained from a 2D Haar wavelet decomposition. Given an input image $x$, the high-frequency representation is computed as $x_{HF} = W_{LH}(x) + W_{HL}(x) + W_{HH}(x)$ where $W_{LH}$, $W_{HL}$, and $W_{HH}$ correspond to detail coefficients that encode sharp edges, stroke boundaries, and fine contour variations. The high-frequency adversarial loss adopts the same hinge loss formulation as the standard adversarial loss, but is applied to $x_{HF}$:
\begin{equation} 
\mathcal{L}_{HF} = \mathbb{E}_{x \sim \mathbf{X}} \left[ \max(0, 1 - D_{HF}(x_{HF})) \right] + \mathbb{E}_{\tilde{y} \sim \mathcal{C},\, z} \left[ \max(0, 1 + D_{HF}(G_{HF}(\tilde{y}, z))) \right],
\end{equation}
where $G_{HF}(\tilde{y}, z)$ denotes the high-frequency components of the generated image $G(\tilde{y}, z)$ obtained using the same decomposition.

For generator training, $\mathcal{L}_{HF}$ encourages the generator to produce images with realistic textures and clearly defined character edges. In our dual-discriminator framework, $\mathcal{L}_{adv}$ provides global structural supervision, while $\mathcal{L}_{HF}$ complements it with targeted supervision on fine-grained details, ensuring handwriting outputs are both structurally coherent and texturally authentic.

\subsubsection{Overall Objectives}

Our model is trained via a min-max adversarial game that balances generator and discriminator objectives, along with auxiliary supervision tasks. The complete training process can be summarized as follows.

\paragraph{Discriminator and Auxiliary Modules.} When maximizing the adversarial objective, we update the global discriminator $D$, Recognizer $R$, and Writer Identifier $W$ independently. Their optimization goals are defined as:
\begin{equation}
\mathcal{L}_D = -\mathcal{L}_{adv}, \quad
\mathcal{L}_{D_{HF}} = -\mathcal{L}_{HF}, \quad
\mathcal{L}_R = \mathcal{L}_{R}^D, \quad
\mathcal{L}_W = \mathcal{L}_{W}^D.
\end{equation}

\paragraph{Generator and Style Encoder.} During the minimization phase, the generator $G$ and the style encoder $E$ are jointly optimized to minimize the following composite loss:
\begin{align}
\mathcal{L}_{G,E} = \mathcal{L}_{adv} + \mathcal{L}_{HF} + \lambda_{R} \mathcal{L}_{R}^G + \lambda_{W} \mathcal{L}_{W}^G + \lambda_{style} \mathcal{L}_{style} + \lambda_{kl} \mathcal{L}_{kl} + \lambda_{FDL}\mathcal{L}_{FDL}.
\end{align}
Each $\lambda$ is a weighting coefficient that balances the influence of its corresponding loss term. These terms supervise different aspects of handwriting synthesis, including realism, legibility, style fidelity, and latent space regularization.

\section{Experiments}\label{sec:experiments}
We evaluate the proposed \textsc{FW-GAN} framework through a comprehensive set of experiments designed to assess both generation quality and practical utility. This section first outlines the design of our evaluation procedure and implementation details, followed by comparisons against state-of-the-art baselines, ablation studies, and cross-language generalization tests. We also investigate the model's impact on downstream handwriting recognition tasks and examine its deployment efficiency.

\subsection{Design of Evaluation Procedure} \label{sub:evaluation_procedure}

Progress in handwritten text generation (HTG) has been slowed by the lack of a unified benchmark for performance assessment. The diversity in evaluation setups across different studies makes it difficult to establish consistent comparisons or draw general conclusions. To tackle this, we propose a clear and replicable evaluation scheme tailored to the specific challenges of HTG.

While the proposed evaluation framework is general-purpose and compatible with multiple datasets, we ground our description using the IAM dataset due to its frequent use in the literature. This dataset contains about 62,857 handwritten English word images produced by 500 individual writers. Rather than relying on the original split, where writers are partitioned into training, validation, and test groups, we adopt the revised division widely used in works such as HWT~\cite{bhunia2021handwriting} and VATr~\cite{pippi2023handwritten}, which assigns 339 writers to training and the remaining 161 to testing.

For consistency with prior models, we use the same preprocessed version of the dataset provided by HWT and VATr. In this version, all images are standardized to a height of 32 pixels, with 16 pixels allocated per character in the text label. To handle varying word lengths, images with a width smaller than 128 pixels are padded, while those exceeding 128 pixels are resized to a fixed size of \(32 \times 128\) pixels. To comprehensively evaluate generative performance, we outline five controlled testing scenarios that challenge both lexical and stylistic generalization:

\begin{itemize}
    \item \textbf{Test Set Replication:} In this setting, the evaluation procedure reconstructs the entire test set by generating each target word exactly as it appears in the original test set. For each generated sample, the style is derived from reference images of the same writer containing the same word, enabling an assessment of the model’s ability to faithfully reproduce both content and writer-specific stylistic attributes.
    \item \textbf{IV-S (In-Vocabulary, Seen Style):} Generate words that appear in the training set using handwriting styles from writers the model has encountered during training.
    \item \textbf{IV-U (In-Vocabulary, Unseen Style):} Synthesize known words using styles from previously unseen writers.
    \item \textbf{OOV-S (Out-of-Vocabulary, Seen Style):} Produce words outside the training vocabulary, rendered in familiar writing styles.
    \item \textbf{OOV-U (Out-of-Vocabulary, Unseen Style):} Combine novel vocabulary with entirely new handwriting styles, posing a dual generalization challenge.
\end{itemize}

In constructing the last four evaluation corpora, we prepare two distinct word pools: one containing in-vocabulary words selected from the training data, and another containing out-of-vocabulary words drawn from an external English corpus. For each configuration, the model is used to generate about 25,000 synthetic word images. These outputs are then compared to real samples using quantitative metrics.

To objectively assess the quality of the generated content, we employ both perceptual and stylistic similarity measures. Specifically, we compute the Fréchet Inception Distance (FID) and Kernel Inception Distance (KID) to capture visual fidelity. These metrics are calculated between real and generated image distributions across the different settings.

\subsection{Implementation Details}\label{subsec:setting}

Our model is implemented using the PyTorch framework\footnote{\url{https://pytorch.org/}} and all experiments are conducted on a single NVIDIA Tesla P100 GPU. Training is performed using the Adam optimizer~\cite{kingma2014adam}, with a learning rate initialized at $0.0002$ and momentum parameters set to $(\beta_1, \beta_2) = (0.5, 0.999)$. Starting from the 35\textsuperscript{th} epoch, the learning rate is gradually reduced following a linear decay schedule.

For the FW-GAN model, we apply a fixed weight of $\lambda_{\text{KL}} = 0.0001$ for the KL divergence term and set $\lambda_{\text{FDL}} = 1$ for the frequency distribution loss. The remaining loss coefficients are adaptively tuned throughout training using a gradient balancing strategy, allowing the model to dynamically regulate competing objectives.

\subsection{Benchmarking Against State-of-the-Art Methods}

To assess the effectiveness of our proposed model, we conduct a comparative analysis with several leading approaches in style-conditioned handwritten text generation. Specifically, we include the Transformer-based Handwriting Transformer (HWT)~\cite{bhunia2021handwriting}, VATr~\cite{pippi2023handwritten}, and the more recent HiGAN~\cite{gan2021higan} and HiGAN+~\cite{gan2022higan+} in our evaluation.

To maintain a fair and consistent experimental setup, we adopt the following protocol: for methods such as HWT and VATr that offer publicly available pretrained models, we utilize their released checkpoints trained on the same dataset splits. For other baselines like HiGAN and HiGAN+, we re-implement and train them from scratch using the identical data partitions employed in HWT and VATr. This ensures that all models are evaluated under the same conditions, eliminating any discrepancies due to differing preprocessing or training schemes. We report quantitative comparisons using two key metrics: Fréchet Inception Distance (FID) and Kernel Inception Distance (KID), as shown in table~\ref{tab:FID-KID} and \ref{tab:oov}. Across all test scenarios, our model (FW-GAN) consistently achieves superior results in terms of FID and KID, indicating both higher visual fidelity and stylistic coherence compared to existing methods. 

Table~\ref{tab:FID-KID} shows that our model achieves the lowest scores on both evaluation metrics, with a FID of 6.530 and KID of 0.20, indicating the highest similarity to real handwriting among all compared methods. In contrast, HiGAN and HiGAN+ report considerably higher FID scores of 17.086 and 16.114, respectively, and KID scores of 1.18 and 0.81, showing a clear performance gap. Transformer-based models, such as HWT (FID: 13.615, KID: 0.49) and VATr (FID: 13.577, KID: 0.47), perform better than earlier methods but still fall short of our results. Notably, our FID is less than half that of the next best-performing model (VATr), and our KID is similarly reduced by more than 50\%.

Table~\ref{tab:oov} reports FID scores across four evaluation scenarios designed to assess both lexical generalization and stylistic robustness. Our model achieves the lowest FID in all four conditions: 24.92 (IV-S), 28.04 (IV-U), 25.46 (OOV-S), and 28.80 (OOV-U). In the IV-S setting, our method improves over the next best model, HWT (26.46), by 1.54 points, and outperforms HiGAN (36.02) by more than 11 points. In IV-U, we surpass VATr (29.94) by 1.90 points, while achieving over 9-point gains compared to HiGAN variants, demonstrating strong generalization of style to unseen writers while retaining lexical familiarity. In the OOV-S scenario, where new words are synthesized in seen styles, our approach leads by 1.01 points over HWT and 1.36 points over VATr. Most notably, in the most challenging OOV-U setting—where both lexical content and style are unseen—our method records 28.80, outperforming VATr (29.50) by 0.70 points, HWT (29.68) by 0.88 points, and HiGAN+ (37.75) by nearly 9 points. While HWT and VATr remain competitive in some OOV cases, neither achieves consistent superiority across all settings. In contrast, our method maintains stable and low FID scores under all conditions, underscoring its robustness to domain shifts and its effectiveness in producing high-fidelity handwriting from familiar contexts to the most challenging OOV-U scenario.

These findings underscore the strength of our proposed model in handling real-world handwriting generation tasks where both the text content and the writer’s style may vary significantly. The ability to generalize under such conditions is essential for practical deployments, and our results reflect a robust and adaptive solution for style-driven handwritten text generation.

\begin{table}[!t]
\centering
\caption{Quantitative comparison of handwriting synthesis quality using FID and KID metrics. Lower values indicate higher similarity to real handwriting.}
\label{tab:FID-KID}
\begin{tabular}{lcc}
\toprule
\textbf{Method} & \textbf{FID} & \textbf{KID} \\
\midrule
HiGAN~\cite{gan2021higan}             & 17.086 & 1.18 \\
HiGAN+~\cite{gan2022higan+}           & 16.114 & 0.81 \\
HWT~\cite{bhunia2021handwriting}        & 13.615 & 0.49  \\
VATr~\cite{pippi2023handwritten}               & 13.577 & 0.47 \\
\textbf{Ours}                  & \textbf{6.530} & \textbf{0.20} \\
\bottomrule
\end{tabular}
\end{table}

\begin{table}[!t]
\centering
\caption{FID scores under different evaluation conditions: In/Out-of-Vocabulary and Seen/Unseen styles (lower is better). Each condition includes 25,000 generated samples.}
\label{tab:oov}
\begin{tabular}{lcccc}
\toprule
\textbf{Method} & \textbf{IV-S} & \textbf{IV-U} & \textbf{OOV-S} & \textbf{OOV-U} \\
\midrule
HiGAN~\cite{gan2021higan}             & 36.02 & 40.94& 36.54 & 41.35  \\
HiGAN+~\cite{gan2022higan+}           & 34.90 & 37.78 & 34.66 & 37.75 \\
HWT~\cite{bhunia2021handwriting}        & 26.46 & 29.86 & 26.47 & 29.68 \\
VATr~\cite{pippi2023handwritten}               & 26.73 & 29.94 & 26.82 & 29.50 \\
\textbf{Ours}                  & \textbf{24.92} & \textbf{28.04}  & \textbf{25.46} & \textbf{28.80} \\
\bottomrule
\end{tabular}
\end{table}
\subsection{Enhancing HTR Performance via Synthetic Data}

To further validate the practical utility of our handwriting synthesis model, we examine its impact on downstream Handwritten Text Recognition (HTR) performance, particularly in scenarios where real training data is scarce. Synthetic handwriting generation aims not only to produce realistic images but also to support recognition tasks by serving as effective training data. 

To simulate a low-resource setting, we begin by training a baseline HTR model using only 5,000 real handwriting samples from the IAM dataset, which the model has never seen before. For this experiment, we adopt a Transformer-based OCR architecture following the TrOCR framework~\cite{Li_2023TrOCR}. We then expand the training dataset by incorporating 25,000 synthetic samples generated by each competing model, including ours, generated from the 5,000 real images with a random lexicon. This augmented dataset (5,000 real + 25,000 synthetic images) is used to retrain the HTR model. By keeping the real data fixed and varying only the synthetic augmentation source, we can directly evaluate the contribution of each synthesis method to recognition performance. We assess HTR effectiveness using three widely accepted metrics:
\begin{itemize}
    \item \textbf{Character Error Rate (CER):} Measures the proportion of character-level errors—substitutions, insertions, and deletions—relative to the total number of characters in the ground truth.
    \item \textbf{Word Error Rate (WER):} Computes the proportion of incorrectly predicted words, capturing recognition quality at the word level.
    \item \textbf{Normalized Edit Distance (NED):} Quantifies the normalized difference between predicted and target sequences; lower values indicate higher transcription fidelity.
\end{itemize}

As shown in Table~\ref{HTR}, all synthetic augmentation methods yield substantial performance gains over the baseline trained with only 5,000 real images, demonstrating the effectiveness of synthetic handwriting in compensating for data scarcity. Among competing approaches, our method achieves the lowest error rates across all three metrics, with a CER of 10.23, NED of 10.09, and WER of 28.18. Compared to the best-performing baseline competitor (VATr), our method reduces CER by 0.39 points, NED by 0.32 points, and WER by 1.47 points. These improvements are consistent across both character- and word-level metrics, indicating that our generated samples not only enhance fine-grained recognition accuracy but also improve overall lexical prediction. The relatively large WER reduction suggests that our model produces more semantically coherent handwriting, enabling the HTR model to better capture entire word structures rather than just isolated characters. This performance consistency across multiple metrics reinforces that our synthesis approach generates data that is both visually realistic and linguistically effective for training robust recognition models.

\begin{table}[!t]
\centering
\caption{HTR performance on the IAM dataset using 5,000 real images and 25,000 synthetic images from each method. Lower scores indicate better performance.}
\label{HTR}
\begin{tabular}{lccc}
\toprule
\textbf{Method} & \textbf{CER} & \textbf{NED} & \textbf{WER} \\
\midrule
5000 real images    & 12.25 & 11.95 & 32.81 \\
HiGAN               & 10.88 & 10.48 & 29.28 \\
HiGAN+              & 10.67 & 10.51 & 29.65 \\
HWT                 & 10.65 & 10.51 & 29.71 \\
VATr                & 10.62 & 10.41 & 29.65 \\
\textbf{Ours}       & \textbf{10.23} & \textbf{10.09} & \textbf{28.18} \\
\bottomrule
\end{tabular}
\end{table}

\subsection{Ablation Study}

To evaluate the contribution of each component in our proposed \textsc{FW-GAN} framework, we conduct an ablation study on the IAM dataset. We start with a baseline configuration (A), which uses only a BigGAN generator backbone, and incrementally add components: Frequency Distribution Loss (FDL), Wave-Modulated Generator, and High-Frequency Discriminator. We assess their impact on generation quality using Fréchet Inception Distance (FID) and downstream recognition performance using Character Error Rate (CER), Normalized Edit Distance (NED), and Word Error Rate (WER).

Table~\ref{tab:ablation-study} summarizes the results. The baseline (A) achieves an FID of 15.94, with CER of 10.57, NED of 10.40, and WER of 29.34. Adding FDL in configuration (B) reduces FID to 10.20, indicating enhanced visual realism, but slightly increases CER (10.74) and NED (10.50), suggesting a minor trade-off in character-level fidelity. Replacing the generator with our Wave-Modulated architecture in configuration (C) further lowers FID to 6.89 and improves recognition performance (CER: 10.32, NED: 10.12, WER: 28.81), demonstrating benefits to both visual quality and text accuracy. Finally, incorporating the High-Frequency Discriminator in configuration (D) yields the best results, with an FID of 6.53, CER of 10.23, NED of 10.09, and WER of 28.18, highlighting its role in capturing fine-grained details that enhance both realism and recognizability.

\begin{figure}[!t]
\centering
\includegraphics[width=1\linewidth]{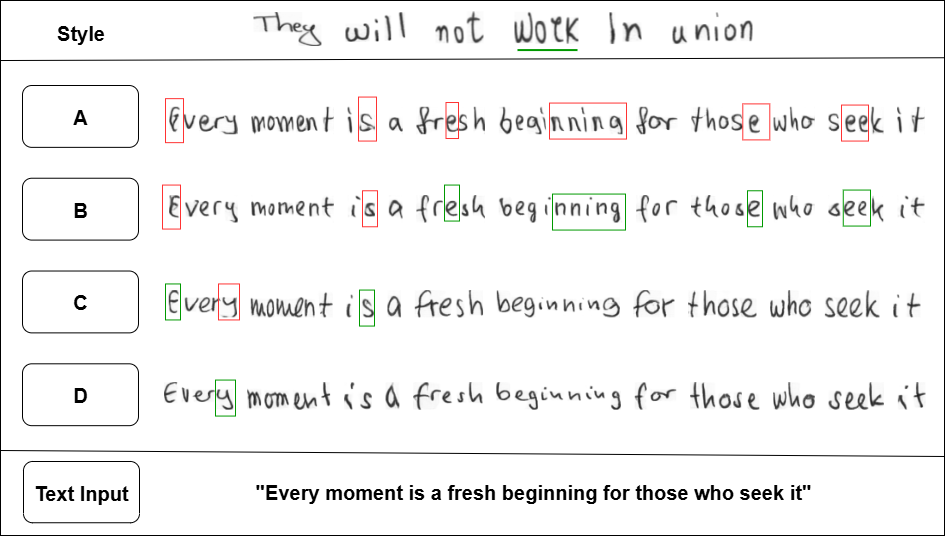}
\caption{Visualization of handwriting images generated by each architectural component, using a single reference image ('work') to generate a sentence. Red boxes highlight failure cases with blurry or faint text; green boxes indicate improved text clarity, stroke style, and color fidelity compared to prior configurations. Models A, B, C, and D correspond to configurations in Table~\ref{tab:ablation-study}.}
\label{fig:ablation_study_visualization}
\end{figure}

To further analyze these improvements, we visualize the generated handwriting images in Figure~\ref{fig:ablation_study_visualization}. In configuration (A), using only the BigGAN generator backbone, the generated text often appears blurry or faint, particularly in stroke edges (highlighted in red boxes), due to the model’s limited ability to capture fine-grained details. Introducing the Frequency Distribution Loss (FDL) in configuration (B) reduces blurriness and enhances text clarity by aligning the spectral characteristics of generated images with real samples, leveraging high-frequency feature distributions to guide the generator toward sharper stroke details. However, stroke color and style still deviate from the reference, indicating incomplete style capture. Incorporating the Wave-Modulated Generator in configuration (C) significantly improves stroke fidelity and color consistency (highlighted in green boxes) by using phase-aware Wave-MLP modules to capture stylistic patterns, though the tail of the character 'y' occasionally remains faint. Finally, adding the High-Frequency Discriminator in configuration (D) produces text with sharp, clear strokes and colors closely matching the reference style, fully eliminating blurriness, including in the character `y'. This is achieved by leveraging wavelet-decomposed high-frequency components to detect and correct subtle artifacts, ensuring precise stroke rendering and stylistic fidelity.
\begin{table}[!t]
\centering
\caption{Ablation study evaluating the effect of each architectural component on generation and recognition performance (IAM dataset). Lower values indicate better performance.}
\label{tab:ablation-study}
\begin{tabular}{lcccc}
\toprule
\textbf{Model} & \textbf{FID} & \textbf{CER} & \textbf{NED} & \textbf{WER} \\
\midrule
Base (BigGAN backbone) (A) & 15.94 & 10.57 & 10.40 & 29.34 \\
(A) + Frequency Distribution Loss (B) & 10.20 & 10.74 & 10.50 & 29.05 \\
(B) + Wave-Modulated Generator (C) & 6.89 & 10.32 & 10.12 & 28.81 \\
(C) + High-Frequency Discriminator (D) & \textbf{6.53} & \textbf{10.23} & \textbf{10.09} & \textbf{28.18} \\
\bottomrule
\end{tabular}
\end{table}

\subsection{Cross-Language Generalization: Vietnamese Handwriting}

To evaluate the generalization ability and robustness of \textsc{FW-GAN} across different writing systems, we extend our experiments beyond English and apply the same experimental setup described in~\ref{subsec:setting} to the Vietnamese language. We use the HANDS-VNOnDB dataset~\cite{HANDS-VNOnDB}, which contains 7,296 handwritten text lines comprising over 480,000 strokes and more than 380,000 characters. Following the dataset’s standard split, we train on handwriting from 106 writers and test on 34 unseen writers.

As reported in Table~\ref{tab:FID-KID-Vietnamese}, our method achieves the lowest FID (5.61) and KID (0.30) among all evaluated approaches. Compared to the best-performing baseline, HWT, \textsc{FW-GAN} reduces FID by 4.24 points and lowers KID from 0.72 to 0.30. These improvements are consistent across both metrics, indicating that our model produces handwriting that is visually closer to real samples while also matching the underlying feature distribution more accurately. Notably, Vietnamese handwriting introduces challenges absent in English, including extensive use of diacritical marks, compound vowels, and tonal variations, all of which increase spatial complexity and stylistic diversity. Despite these complexities, \textsc{FW-GAN} successfully captures fine-grained structural and stylistic cues, producing images that are both visually convincing and stylistically coherent across writers. This strong performance on a typologically different language demonstrates the model’s robustness in cross-language scenarios, confirming its ability to generalize high-fidelity synthesis beyond the training language.

\begin{table}[!t]
\centering
\caption{Quantitative comparison of handwriting synthesis quality using FID and KID metrics on HANDS-VNOnDB. Lower values indicate higher similarity to real handwriting.}
\label{tab:FID-KID-Vietnamese}
\begin{tabular}{lcc}
\toprule
\textbf{Method} & \textbf{FID} & \textbf{KID} \\
\midrule
VATr~\cite{pippi2023handwritten}       & 23.88 & 2.72 \\
HiGAN~\cite{gan2021higan}              & 11.26 & 0.79 \\
HWT~\cite{bhunia2021handwriting}       & 9.85 & 0.72 \\
\textbf{Ours}                       & \textbf{5.61} & \textbf{0.30} \\
\bottomrule
\end{tabular}
\end{table}

\subsection{Model Size and Deployment Efficiency}

We assess the parameter efficiency of our proposed model by comparing its memory footprint against representative handwriting synthesis baselines, as detailed in~\ref{tab:size}. Our model exhibits a total size of just 52.96 MB, placing it among the most compact solutions—substantially smaller than HWT (131.3 MB), VATr (155.72 MB), and only marginally larger than HiGAN+ (21.7 MB). The compact nature of our architecture makes it highly suitable for real-world use cases where computational resources and memory are limited. Despite its smaller size, our model maintains high-quality output and strong recognition compatibility, offering a favorable trade-off between efficiency and performance.

\begin{table}[!t]
\centering
\caption{Comparison of model sizes (in megabytes) for handwriting synthesis. Only the generator (Gen) and encoder (Enc) modules are considered.}
\label{tab:size}
\begin{tabular}{lccc}
\toprule
\textbf{Method} & \textbf{Gen (MB)} & \textbf{Enc (MB)} & \textbf{Total (MB)} \\
\midrule
HWT~\cite{bhunia2021handwriting}      & 80.7 & 50.6 & 131.3 \\
VATr~\cite{pippi2023handwritten}      & 113.11 & 42.61 & 155.72 \\
HiGAN~\cite{gan2021higan}             & 38.6 & 20.5 & 59.1 \\
HiGAN+~\cite{gan2022higan+}           & \textbf{15.0} & 6.7 & \textbf{21.7} \\
\textbf{Ours}                         & 46.37 & \textbf{6.59} & 52.96 \\
\bottomrule
\end{tabular}
\end{table}

\subsection{Visual Comparison}
In addition to the quantitative evaluations presented in previous sections, we conduct a qualitative analysis to further assess the visual realism and style preservation capability of \textsc{FW-GAN}. While metrics such as FID, KID, and recognition-based scores provide objective measures of performance, they do not fully capture subtle handwriting characteristics such as stroke continuity, letter slant, or spacing regularity. To address this, we visually compare our model with several state-of-the-art baselines under two complementary settings: generation quality, where models synthesize handwriting given a reference style and new content, and reconstruction fidelity, where models aim to replicate a target style as closely as possible. The following subsections present side-by-side examples highlighting key visual differences between methods.

\begin{figure}[!t]
\centering
\includegraphics[width=1\linewidth]{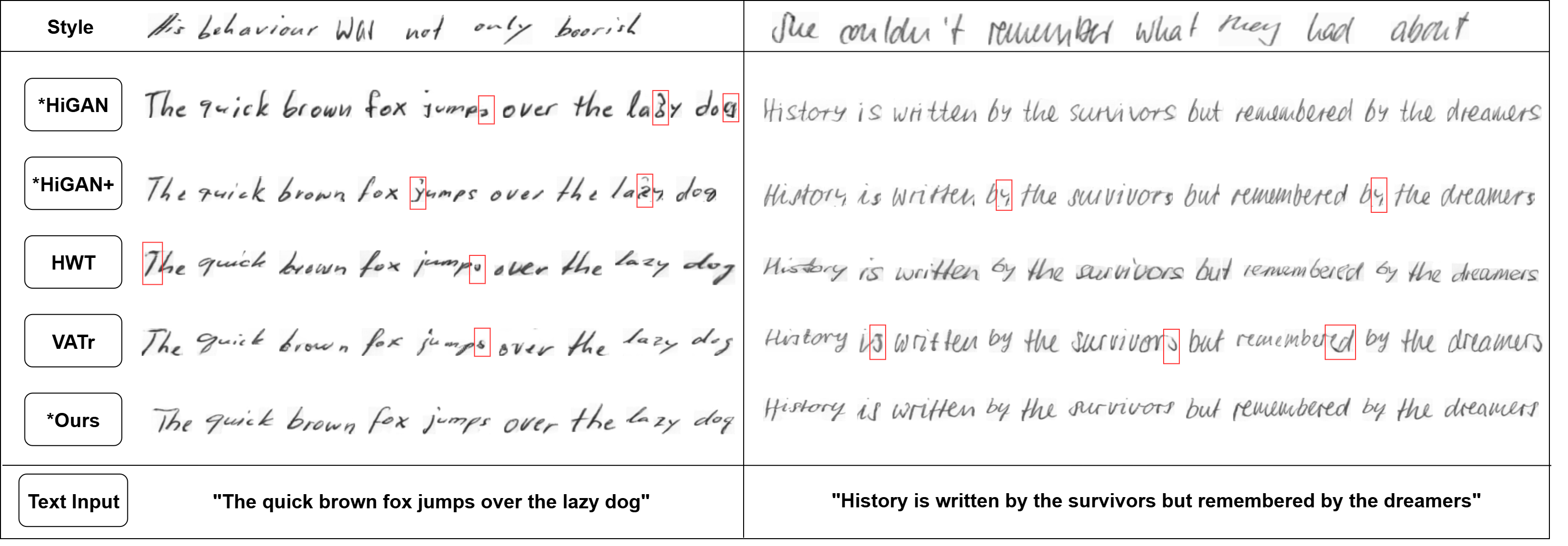}
\caption{Qualitative comparison of handwriting generated from the IAM dataset across models, using identical style and content inputs. Each row corresponds to a different model, with red boxes highlighting failure cases (e.g., blurry or incorrect characters). The asterisk (*) denotes a one-shot model requiring only a single reference image.}
\label{fig:qualitative_gen_eng}
\end{figure}

\begin{figure}[!t]
\centering
\includegraphics[width=1\linewidth]{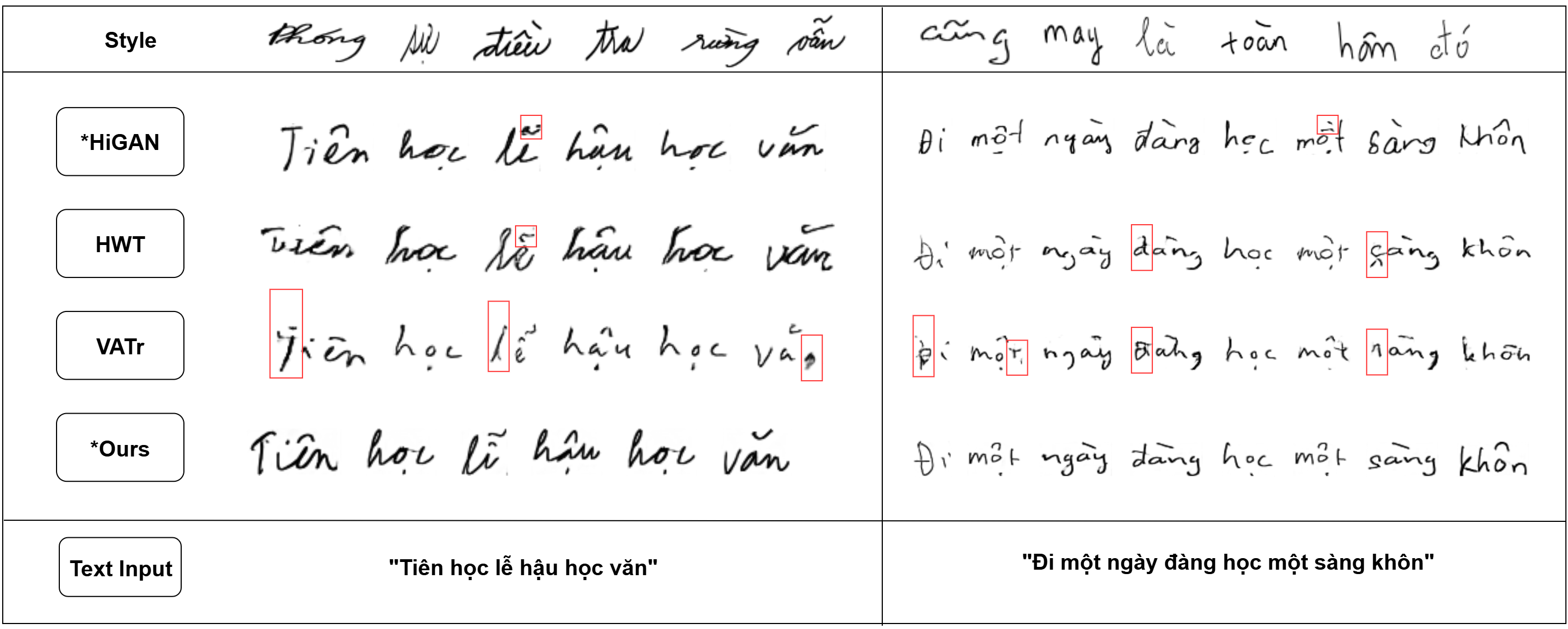}
\caption{Qualitative comparison of handwriting generated from the HANDS-VNOnDB dataset across models, using identical style and content inputs. Each row corresponds to a different model, with red boxes highlighting failure cases (e.g., blurry or incorrect characters). The asterisk (*) denotes a one-shot model requiring only a single reference image.}
\label{fig:qualitative_gen_vietnamese}
\end{figure}

\subsubsection{Generation Quality}
Figures~\ref{fig:qualitative_gen_eng} and \ref{fig:qualitative_gen_vietnamese} present qualitative comparisons of handwriting generated by our \textsc{FW-GAN} model against baselines, including HiGAN~\cite{gan2021higan}, HiGAN+~\cite{gan2022higan+}, HWT~\cite{bhunia2021handwriting}, and VATr~\cite{pippi2023handwritten}, on the English (IAM) and Vietnamese (HANDS-VNOnDB) datasets. All models were conditioned on identical style references (top row) and content inputs (bottom row). Our method consistently produces stylistically coherent and visually clear handwriting across both datasets.

For the IAM dataset, \textsc{FW-GAN} generates text with smooth stroke continuity, accurate stroke thickness, and color fidelity, closely matching the reference style. Competing models, such as HiGAN and VATr, exhibit character errors (e.g., blurry or distorted 's' and 'z', highlighted in red boxes) and inconsistent color or shading. In contrast, our model’s phase-aware Wave-MLP and frequency-guided components ensure legible characters with precise stylistic attributes. For the HANDS-VNOnDB dataset, where complex character shapes and diacritics pose additional challenges, \textsc{FW-GAN} significantly outperforms baselines. While VATr and HWT produce broken strokes and diacritic misplacements (highlighted in red boxes), our model leverages its Frequency Distribution Loss and High-Frequency Discriminator to generate sharp, coherent text with accurate slant, hue, and stroke thickness, demonstrating robust generalization across languages.

\begin{figure}[!t]
\centering
\includegraphics[width=1\linewidth]{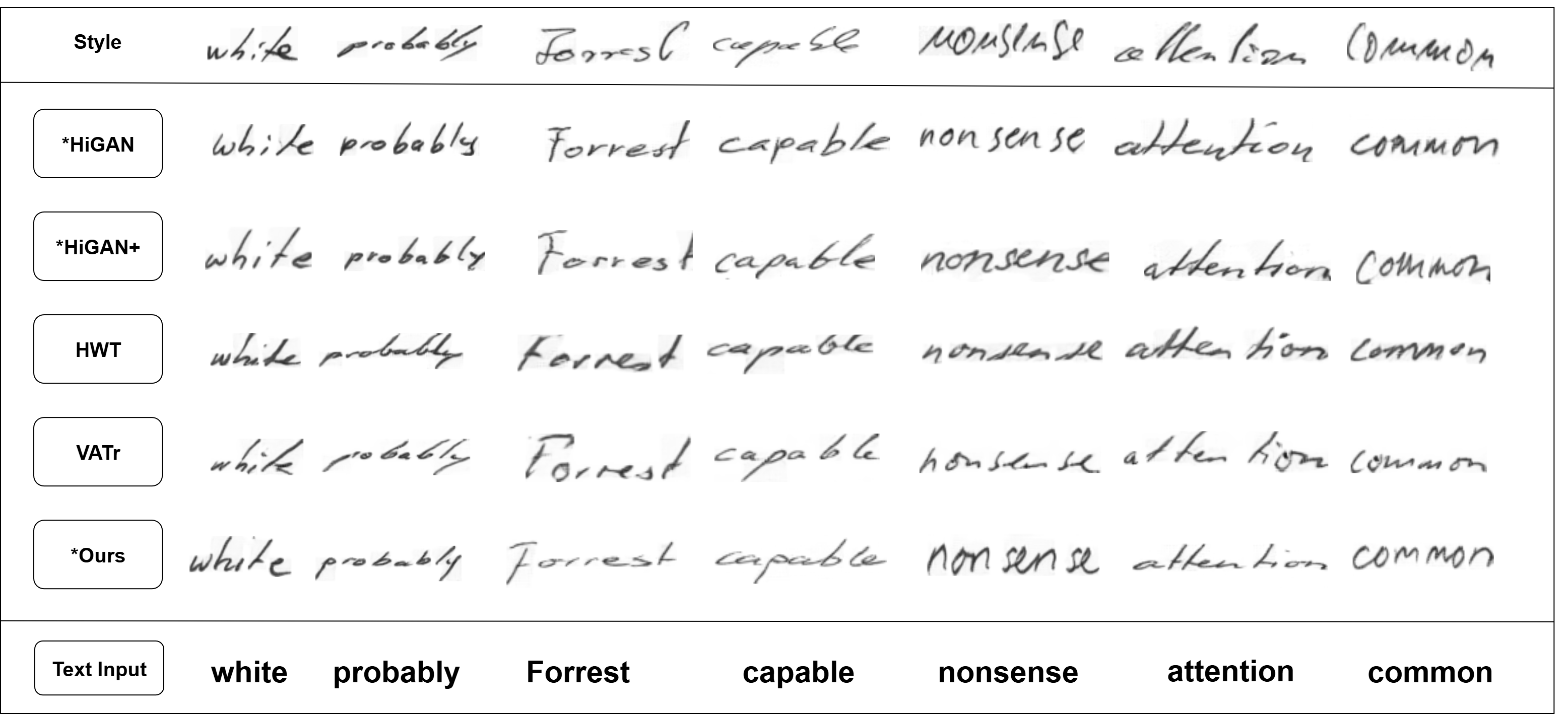}
\caption{Qualitative reconstruction results. Each row corresponds to a different model, conditioned on the same ground-truth style (top row) and target text (bottom row). The goal is to reproduce the target content while preserving the handwriting style.}
\label{fig:reconstruction}
\end{figure}

\subsubsection{Reconstruction Fidelity}
Figure~\ref{fig:reconstruction} presents reconstruction results where each model is given ground-truth style images along with a target content prompt, with the goal of reproducing handwriting that closely matches the reference style.

Our model preserves character size, color, and brightness with high fidelity, while capturing the handwriting slant more accurately than others. It also avoids common issues such as stroke omission, for instance, in HiGAN’s rendering of the letter “t” in “white,” where the crossbar is missing. In comparison, HWT often produces overly bold characters (e.g., in “Forrest” or “attention”), whereas VATr tends to generate strokes that appear less cohesive and somewhat fragmented.

By contrast, our method consistently produces smooth, well-connected letters that align with the reference style in both color tone and stroke intensity. While HiGAN and HiGAN+ deliver relatively strong performance, our approach more faithfully reproduces subtle stylistic nuances, such as slight variations in stroke darkness and character shapes. These results demonstrate the effectiveness of our Frequency-Driven and Wave-Modulated MLP Generator in capturing fine-grained handwriting details.

\section{Conclusion}\label{sec:conclusion}
In this paper, we introduce \textsc{FW-GAN}, a novel GAN-based model designed for one-shot handwriting synthesis, addressing the challenge of generating realistic, style-consistent handwritten text from a single sample. By integrating phase-aware Wave-MLP modules into the generator, our model captures intricate calligraphic styles, overcoming the limitations of conventional convolutional architectures. A frequency-guided discriminator, paired with a novel Frequency Distribution Loss, aligns the spectral characteristics of generated samples with authentic handwriting, enhancing fine-grained details such as stroke edges and pen pressure variations. To the best of our knowledge, this work represents one of the first explorations of frequency-aware mechanisms and MLP-augmented architectures in handwriting synthesis. The proposed framework not only advances the field of generative modeling for handwriting but also provides a robust tool for data augmentation in HTR pipelines and personalized text generation.

\bibliographystyle{elsarticle-num}
\bibliography{references} % Replace 'references' with the name of your .bib file (without .bib extension)

\begin{thebibliography}{10}
\expandafter\ifx\csname url\endcsname\relax
  \def\url#1{\texttt{#1}}\fi
\expandafter\ifx\csname urlprefix\endcsname\relax\def\urlprefix{URL }\fi
\expandafter\ifx\csname href\endcsname\relax
  \def\href#1#2{#2} \def\path#1{#1}\fi

\bibitem{HANDS-VNOnDB}
H.~T. Nguyen, C.~T. Nguyen, P.~T. Bao, M.~Nakagawa,
  \href{https://www.sciencedirect.com/science/article/pii/S0031320318300141}{A
  database of unconstrained vietnamese online handwriting and recognition
  experiments by recurrent neural networks}, Pattern Recognition 78 (2018)
  291--306.
\newblock \href {https://doi.org/https://doi.org/10.1016/j.patcog.2018.01.013}
  {\path{doi:https://doi.org/10.1016/j.patcog.2018.01.013}}.
\newline\urlprefix\url{https://www.sciencedirect.com/science/article/pii/S0031320318300141}

\bibitem{kleber2013cvl}
F.~Kleber, S.~Fiel, M.~Diem, R.~Sablatnig, {CVL-DataBase: An Off-Line Database
  for Writer Retrieval, Writer Identification and Word Spotting}, in: ICDAR,
  2013.

\bibitem{ICFHR2018}
I.~Pratikakis, K.~Zagori, P.~Kaddas, B.~Gatos, Icfhr 2018 competition on
  handwritten document image binarization (h-dibco 2018), in: 2018 16th
  International Conference on Frontiers in Handwriting Recognition (ICFHR),
  2018, pp. 489--493.
\newblock \href {https://doi.org/10.1109/ICFHR-2018.2018.00091}
  {\path{doi:10.1109/ICFHR-2018.2018.00091}}.

\bibitem{NabucoLatin}
R.~D. Lins, \href{https://api.semanticscholar.org/CorpusID:2896293}{Nabuco -
  two decades of document processing in latin america}, J. Univers. Comput.
  Sci. 17 (2011) 151--161.
\newline\urlprefix\url{https://api.semanticscholar.org/CorpusID:2896293}

\bibitem{graves2013generating}
A.~Graves, {Generating Sequences with Recurrent Neural Networks}, arXiv
  preprint arXiv:1308.0850 (2013).

\bibitem{bhunia2021handwriting}
A.~K. Bhunia, S.~Khan, H.~Cholakkal, R.~M. Anwer, F.~S. Khan, M.~Shah,
  {Handwriting Transformers}, in: ICCV, 2021.

\bibitem{Zhi_2020FPrinciple}
Z.-Q. J.~X. Zhi-Qin John~Xu, Y.~Z. Yaoyu~Zhang, T.~L. Tao~Luo, Y.~X.
  Yanyang~Xiao, Z.~M. Zheng~Ma,
  \href{http://dx.doi.org/10.4208/cicp.OA-2020-0085}{Frequency principle:
  Fourier analysis sheds light on deep neural networks}, Communications in
  Computational Physics 28~(5) (2020) 1746–1767.
\newblock \href {https://doi.org/10.4208/cicp.oa-2020-0085}
  {\path{doi:10.4208/cicp.oa-2020-0085}}.
\newline\urlprefix\url{http://dx.doi.org/10.4208/cicp.OA-2020-0085}

\bibitem{Tang_2022WaveMlp}
Y.~Tang, K.~Han, J.~Guo, C.~Xu, Y.~Li, C.~Xu, Y.~Wang,
  \href{http://dx.doi.org/10.1109/CVPR52688.2022.01066}{An image patch is a
  wave: Phase-aware vision mlp}, in: 2022 IEEE/CVF Conference on Computer
  Vision and Pattern Recognition (CVPR), IEEE, 2022, p. 10925–10934.
\newblock \href {https://doi.org/10.1109/cvpr52688.2022.01066}
  {\path{doi:10.1109/cvpr52688.2022.01066}}.
\newline\urlprefix\url{http://dx.doi.org/10.1109/CVPR52688.2022.01066}

\bibitem{Ni_2024FDL}
Z.~Ni, J.~Wu, Z.~Wang, W.~Yang, H.~Wang, L.~Ma,
  \href{http://dx.doi.org/10.1109/CVPR52733.2024.00281}{Misalignment-robust
  frequency distribution loss for image transformation}, in: 2024 IEEE/CVF
  Conference on Computer Vision and Pattern Recognition (CVPR), IEEE, 2024, p.
  2910–2919.
\newblock \href {https://doi.org/10.1109/cvpr52733.2024.00281}
  {\path{doi:10.1109/cvpr52733.2024.00281}}.
\newline\urlprefix\url{http://dx.doi.org/10.1109/CVPR52733.2024.00281}

\bibitem{aksan2018deepwriting}
E.~Aksan, F.~Pece, O.~Hilliges, {DeepWriting: Making digital ink editable via
  deep generative modeling}, in: CHI, ACM, 2018.

\bibitem{aksan2018stcn}
E.~Aksan, O.~Hilliges, {STCN: Stochastic Temporal Convolutional Networks}, in:
  ICLR, 2018.

\bibitem{kotani2020generating}
A.~Kotani, S.~Tellex, J.~Tompkin, {Generating Handwriting via Decoupled Style
  Descriptors}, in: ECCV, 2020.

\bibitem{ji2019generative}
B.~Ji, T.~Chen, {Generative Adversarial Network for Handwritten Text}, arXiv
  preprint arXiv:1907.11845 (2019).

\bibitem{wang2005combining}
J.~Wang, C.~Wu, Y.-Q. Xu, H.-Y. Shum, {Combining Shape and Physical Models for
  On-line Cursive Handwriting Synthesis}, IJDAR 7~(4) (2005) 219--227.

\bibitem{lin2007style}
Z.~Lin, L.~Wan, Style-preserving english handwriting synthesis, Pattern
  Recognit. 40~(7) (2007) 2097--2109.

\bibitem{thomas2009synthetic}
A.~O. Thomas, A.~Rusu, V.~Govindaraju, {Synthetic Handwritten CAPTCHAs},
  Pattern Recognit. 42~(12) (2009) 3365--3373.

\bibitem{haines2016my}
T.~Haines, O.~Mac~Aodha, G.~Brostow, {My Text in Your Handwriting}, {ACM Trans.
  Graphics} 35~(3) (2016).

\bibitem{alonso2019adversarial}
E.~Alonso, B.~Moysset, R.~Messina, {Adversarial Generation of Handwritten Text
  Images Conditioned on Sequences}, in: ICDAR, IEEE Computer Society, 2019.

\bibitem{fogel2020scrabblegan}
S.~Fogel, H.~Averbuch-Elor, S.~Cohen, S.~Mazor, R.~Litman, {ScrabbleGAN:
  Semi-Supervised Varying Length Handwritten Text Generation}, in: CVPR, 2020.

\bibitem{kang2020ganwriting}
L.~Kang, P.~Riba, Y.~Wang, M.~Rusi{\~n}ol, A.~Forn{\'e}s, M.~Villegas,
  {GANwriting: Content-Conditioned Generation of Styled Handwritten Word
  Images}, in: ECCV, 2020.

\bibitem{gan2021higan}
J.~Gan, W.~Wang, {HiGAN: Handwriting Imitation Conditioned on Arbitrary-Length
  Texts and Disentangled Styles}, in: AAAI, 2021.

\bibitem{mattick2021smartpatch}
A.~Mattick, M.~Mayr, M.~Seuret, A.~Maier, V.~Christlein, {SmartPatch: Improving
  Handwritten Word Imitation with Patch Discriminators}, in: ICDAR, 2021.

\bibitem{luo2022slogan}
C.~Luo, Y.~Zhu, L.~Jin, Z.~Li, D.~Peng, {SLOGAN: Handwriting Style Synthesis
  for Arbitrary-Length and Out-of-Vocabulary Text}, IEEE Trans. Neural Netw.
  Learn. Syst. (2022).

\bibitem{krishnan2021textstylebrush}
P.~Krishnan, R.~Kovvuri, G.~Pang, B.~Vassilev, T.~Hassner, {TextStyleBrush:
  Transfer of Text Aesthetics from a Single Example}, arXiv e-prints (2021)
  arXiv--2106.

\bibitem{davis2020text}
B.~Davis, C.~Tensmeyer, B.~Price, C.~Wigington, B.~Morse, R.~Jain, {Text and
  Style Conditioned GAN for Generation of Offline Handwriting Lines}, in: BMVC,
  2020.

\bibitem{brock2019biggan}
A.~Brock, J.~Donahue, K.~Simonyan, {Large Scale {GAN} Training for High
  Fidelity Natural Image Synthesis}, in: ICLR, 2019.

\bibitem{pippi2023handwritten}
V.~Pippi, S.~Cascianelli, R.~Cucchiara, {Handwritten Text Generation from
  Visual Archetypes}, in: CVPR, 2023.

\bibitem{tolstikhin2021mlpmixer}
I.~Tolstikhin, N.~Houlsby, A.~Kolesnikov, L.~Beyer, X.~Zhai, T.~Unterthiner,
  J.~Yung, A.~Steiner, D.~Keysers, J.~Uszkoreit, M.~Lucic, A.~Dosovitskiy,
  Mlp-mixer: An all-mlp architecture for vision (2021).
\newblock \href {http://arxiv.org/abs/2105.01601} {\path{arXiv:2105.01601}}.

\bibitem{ding2021repmlp}
X.~Ding, C.~Xia, X.~Zhang, X.~Chu, J.~Han, G.~Ding, Repmlp: Re-parameterizing
  convolutions into fully-connected layers for image recognition (2021).
\newblock \href {http://arxiv.org/abs/2105.01883} {\path{arXiv:2105.01883}}.

\bibitem{Guo_2022BeyondSelfAttention}
M.-H. Guo, Z.-N. Liu, T.-J. Mu, S.-M. Hu,
  \href{http://dx.doi.org/10.1109/TPAMI.2022.3211006}{Beyond self-attention:
  External attention using two linear layers for visual tasks}, IEEE
  Transactions on Pattern Analysis and Machine Intelligence (2022) 1–13\href
  {https://doi.org/10.1109/tpami.2022.3211006}
  {\path{doi:10.1109/tpami.2022.3211006}}.
\newline\urlprefix\url{http://dx.doi.org/10.1109/TPAMI.2022.3211006}

\bibitem{melaskyriazi2021need}
L.~Melas-Kyriazi, Do you even need attention? a stack of feed-forward layers
  does surprisingly well on imagenet (2021).
\newblock \href {http://arxiv.org/abs/2105.02723} {\path{arXiv:2105.02723}}.

\bibitem{liu2021swin}
Z.~Liu, Y.~Lin, Y.~Cao, H.~Hu, Y.~Wei, Z.~Zhang, S.~Lin, B.~Guo, Swin
  transformer: Hierarchical vision transformer using shifted windows (2021).
\newblock \href {http://arxiv.org/abs/2103.14030} {\path{arXiv:2103.14030}}.

\bibitem{ho2019axial}
J.~Ho, N.~Kalchbrenner, D.~Weissenborn, T.~Salimans, Axial attention in
  multidimensional transformers (2019).
\newblock \href {http://arxiv.org/abs/1912.12180} {\path{arXiv:1912.12180}}.

\bibitem{luo2019theory}
T.~Luo, Z.~Ma, Z.-Q.~J. Xu, Y.~Zhang, Theory of the frequency principle for
  general deep neural networks, arXiv preprint arXiv:1906.09235 (2019).

\bibitem{jacot2018ntk}
A.~Jacot, F.~Gabriel, C.~Hongler, Neural tangent kernel: Convergence and
  generalization in neural networks, in: Advances in neural information
  processing systems, 2018, pp. 8571--8580.

\bibitem{luo2020theory}
T.~Luo, Z.~Ma, Z.-Q.~J. Xu, Y.~Zhang,
  \href{http://dx.doi.org/10.1137/21m1444400}{On the exact computation of
  linear frequency principle dynamics and its generalization}, SIAM Journal on
  Mathematics of Data Science 4~(4) (2022) 1272–1292.
\newblock \href {https://doi.org/10.1137/21m1444400}
  {\path{doi:10.1137/21m1444400}}.
\newline\urlprefix\url{http://dx.doi.org/10.1137/21m1444400}

\bibitem{zhang2019explicitizing}
Y.~Zhang, Z.-Q.~J. Xu, T.~Luo, Z.~Ma, Explicitizing an implicit bias of the
  frequency principle in two-layer neural networks (2019).
\newblock \href {http://arxiv.org/abs/1905.10264} {\path{arXiv:1905.10264}}.

\bibitem{bordelon2020spectrum}
B.~Bordelon, A.~Canatar, C.~Pehlevan, Spectrum dependent learning curves in
  kernel regression and wide neural networks (2020).
\newblock \href {http://arxiv.org/abs/2002.02561} {\path{arXiv:2002.02561}}.

\bibitem{cao2019towards}
Y.~Cao, Z.~Fang, Y.~Wu, D.-X. Zhou, Q.~Gu, Towards understanding the spectral
  bias of deep learning, arXiv preprint arXiv:1912.01198 (2019).

\bibitem{basri2019convergence}
R.~Basri, D.~Jacobs, Y.~Kasten, S.~Kritchman, The convergence rate of neural
  networks for learned functions of different frequencies (2019).
\newblock \href {http://arxiv.org/abs/1906.00425} {\path{arXiv:1906.00425}}.

\bibitem{yang2019fine}
G.~Yang, H.~Salman, A fine-grained spectral perspective on neural networks,
  arXiv preprint arXiv:1907.10599 (2019).

\bibitem{e2019frequency}
W.~E, C.~Ma, L.~Wu, Machine learning from a continuous viewpoint, arXiv
  preprint arXiv:1912.12777 (2019).

\bibitem{gan2022higan+}
J.~Gan, W.~Wang, J.~Leng, X.~Gao, {HiGAN+: Handwriting Imitation GAN with
  Disentangled Representations}, {ACM Trans. Graphics} 42~(1) (2022) 1--17.

\bibitem{Braginsky1992_quantummeasurement}
V.~B. Braginsky, F.~Y. Khalili, K.~S. Thorne, Quantum Measurement, Cambridge
  University Press, 1992.

\bibitem{Jacobs2006_quantummeasurement}
K.~Jacobs, D.~A. Steck, \href{http://dx.doi.org/10.1080/00107510601101934}{A
  straightforward introduction to continuous quantum measurement}, Contemporary
  Physics 47~(5) (2006) 279–303.
\newblock \href {https://doi.org/10.1080/00107510601101934}
  {\path{doi:10.1080/00107510601101934}}.
\newline\urlprefix\url{http://dx.doi.org/10.1080/00107510601101934}

\bibitem{lim2017geometric}
J.~H. Lim, J.~C. Ye, {Geometric GAN}, arXiv preprint arXiv:1705.02894 (2017).

\bibitem{Graves2006ctc}
A.~Graves, S.~Fern\'{a}ndez, F.~Gomez, J.~Schmidhuber,
  \href{https://doi.org/10.1145/1143844.1143891}{Connectionist temporal
  classification: labelling unsegmented sequence data with recurrent neural
  networks}, in: Proceedings of the 23rd International Conference on Machine
  Learning, ICML '06, Association for Computing Machinery, New York, NY, USA,
  2006, p. 369–376.
\newblock \href {https://doi.org/10.1145/1143844.1143891}
  {\path{doi:10.1145/1143844.1143891}}.
\newline\urlprefix\url{https://doi.org/10.1145/1143844.1143891}

\bibitem{chen2016infogan}
X.~Chen, Y.~Duan, R.~Houthooft, J.~Schulman, I.~Sutskever, P.~Abbeel, Infogan:
  Interpretable representation learning by information maximizing generative
  adversarial nets (2016).
\newblock \href {http://arxiv.org/abs/1606.03657} {\path{arXiv:1606.03657}}.

\bibitem{zhu2017multimodal}
J.-Y. Zhu, R.~Zhang, D.~Pathak, T.~Darrell, A.~A. Efros, O.~Wang, E.~Shechtman,
  Toward multimodal image-to-image translation (2017).
\newblock \href {http://arxiv.org/abs/1711.11586} {\path{arXiv:1711.11586}}.

\bibitem{lee2018diverse}
H.-Y. Lee, H.-Y. Tseng, J.-B. Huang, M.~Singh, M.-H. Yang,
  \href{http://dx.doi.org/10.1007/978-3-030-01246-5_3}{Diverse Image-to-Image
  Translation via Disentangled Representations}, Springer International
  Publishing, 2018, p. 36–52.
\newblock \href {https://doi.org/10.1007/978-3-030-01246-5_3}
  {\path{doi:10.1007/978-3-030-01246-5_3}}.
\newline\urlprefix\url{http://dx.doi.org/10.1007/978-3-030-01246-5_3}

\bibitem{kingma2014adam}
D.~P. Kingma, J.~Ba, Adam: A method for stochastic optimization, in: ICLR,
  2015.

\bibitem{Li_2023TrOCR}
M.~Li, T.~Lv, J.~Chen, L.~Cui, Y.~Lu, D.~Florencio, C.~Zhang, Z.~Li, F.~Wei,
  \href{http://dx.doi.org/10.1609/aaai.v37i11.26538}{Trocr: Transformer-based
  optical character recognition with pre-trained models}, Proceedings of the
  AAAI Conference on Artificial Intelligence 37~(11) (2023) 13094–13102.
\newblock \href {https://doi.org/10.1609/aaai.v37i11.26538}
  {\path{doi:10.1609/aaai.v37i11.26538}}.
\newline\urlprefix\url{http://dx.doi.org/10.1609/aaai.v37i11.26538}

\end{thebibliography}

\end{document}